\documentclass[%
 aip,
 amsmath,amssymb,
 reprint,%
]{revtex4-1}

\usepackage{graphicx, float}
\usepackage{dcolumn}
\usepackage{bm}

\usepackage[utf8]{inputenc}
\usepackage[T1]{fontenc}
\usepackage{mathptmx}
\usepackage{etoolbox}

\usepackage{multirow}
\usepackage[table,xcdraw]{xcolor}
\usepackage[flushleft]{threeparttable}
\usepackage[varg]{txfonts}

\makeatletter
\def\@email#1#2{%
 \endgroup
 \patchcmd{\titleblock@produce}
  {\frontmatter@RRAPformat}
  {\frontmatter@RRAPformat{\produce@RRAP{*#1\href{mailto:#2}{#2}}}\frontmatter@RRAPformat}
  {}{}
}%
\makeatother
\begin{document}

\preprint{AIP/123-QED}

\title[]{Learning A Simulation-based Visual Policy for \\Real-world Peg In Unseen Holes}

\author{Liang Xie}
 \altaffiliation[Also at ]{the State Key Laboratory of Industrial Control Technology and Institute of Cyber-Systems and Control, Zhejiang University, Zhejiang, China.}
\affiliation{College of Control Science and Engineering, Zhejiang University, Zhejiang, China.}

\author{Hongxiang Yu}
 \altaffiliation[Also at ]{the State Key Laboratory of Industrial Control Technology and Institute of Cyber-Systems and Control, Zhejiang University, Zhejiang, China.}
\affiliation{College of Control Science and Engineering, Zhejiang University, Zhejiang, China.}

\author{Kechun Xu}
 \altaffiliation[Also at ]{the State Key Laboratory of Industrial Control Technology and Institute of Cyber-Systems and Control, Zhejiang University, Zhejiang, China.}
\affiliation{College of Control Science and Engineering, Zhejiang University, Zhejiang, China.}

\author{Tong Yang}
 \altaffiliation[Also at ]{the State Key Laboratory of Industrial Control Technology and Institute of Cyber-Systems and Control, Zhejiang University, Zhejiang, China.}
\affiliation{College of Control Science and Engineering, Zhejiang University, Zhejiang, China.}

\author{Minhang Wang}
 \affiliation{The Application Innovate Lab, Huawei Incorporated Company, China.}

\author{Haojian Lu}
 \altaffiliation[Also at ]{the State Key Laboratory of Industrial Control Technology and Institute of Cyber-Systems and Control, Zhejiang University, Zhejiang, China.}
\affiliation{College of Control Science and Engineering, Zhejiang University, Zhejiang, China.}

\author{Rong Xiong}
 \altaffiliation[Also at ]{the State Key Laboratory of Industrial Control Technology and Institute of Cyber-Systems and Control, Zhejiang University, Zhejiang, China.}
\affiliation{College of Control Science and Engineering, Zhejiang University, Zhejiang, China.}

\author{Yue Wang}
\email{ywang24@zju.edu.cn}
 \altaffiliation[Also at ]{the State Key Laboratory of Industrial Control Technology and Institute of Cyber-Systems and Control, Zhejiang University, Zhejiang, China.}
\affiliation{College of Control Science and Engineering, Zhejiang University, Zhejiang, China.}



\date{\today}


\maketitle


\begin{quotation}
This paper proposes a learning-based visual peg-in-hole that enables training with several shapes in simulation, and adapting to arbitrary unseen shapes in real world with minimal sim-to-real cost. The core idea is to decouple the generalization of the sensory-motor policy to the design of a fast-adaptable perception module and a simulated generic policy module. The framework consists of a segmentation network (SN), a virtual sensor network (VSN), and a controller network (CN). Concretely, the VSN is trained to measure the pose of the unseen shape from a segmented image. After that, given the shape-agnostic pose measurement, the CN is trained to achieve generic peg-in-hole. Finally, when applying to real unseen holes, we only have to fine-tune the SN required by the simulated VSN+CN. To further minimize the transfer cost, we propose to automatically collect and annotate the data for the SN after one-minute human teaching. Simulated and real-world results are presented under the configurations of eye-to/in-hand. An electric vehicle charging system with the proposed policy inside achieves a 10/10 success rate in 2$\sim$3s, using only hundreds of auto-labeled samples for the SN transfer.
\end{quotation}
\section{Introduction}
\begin{figure}
\includegraphics[height=103mm,width=0.48\textwidth]{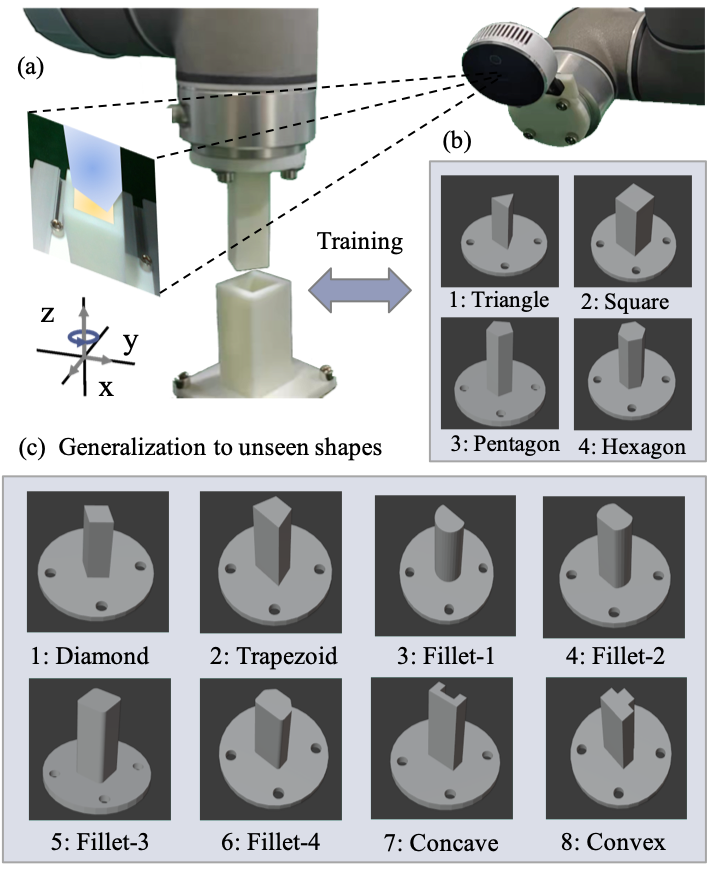}
\caption{Conceptual illustration of the precision peg-in-hole. The peg is mounted on the end-effector of the manipulator whose motion is guided by an image-based sensor, as shown in (a). 
The vision-based solution is particularly preferred in this work, because of its ability to capture the global scene information as well as the semantic and geometric object properties. 
The training process of any algorithm can only be carried out within a limited set of peg shapes, shown in (b), while in practical the shape of the peg to be manipulated may be arbitrary, such as the testing cases shown in (c). 
This work aims to propose a framework for precision peg-in-hole towards easy generalization to arbitrary unseen shapes. 
}
\label{setting}
\end{figure}

Robotic peg-in-hole~\cite{2019Compare} is a routine skill with a wide spectrum of applications ranging from industrial manufacturing to recently human's daily housework, such as placing the key into the lock and inserting the bolt into the nut. In these applications, the shapes of the peg/hole to be manipulated may be arbitrary and unseen in prior, thus developing the capability to deal with unseen shapes becomes non-trivial for all existing peg-in-hole solutions. An expected learning scenario is that the policy is \emph{trained with several shapes, and tested to another set of shapes unseen in the training set}, see Fig.~\ref{setting} for illustration. In this way, the robot can better adapt to the diverse demands, e.g. an electric vehicle (EV) charging robot is able to handle new protocols of charging interface without additional expert programming.

Towards the goal, the existing force-based insertion policies have achieved a high success rate for holes with sub-$mm$ tolerance, given the initial misaligned error of about 1$mm$ \cite{2021Offline}. But it fails when the initial error becomes larger. One recent work \cite{2021Robust} pushes the robustness against the initial error up to 4$mm$ and 4$^{\circ}$, with the cost that the generalization is achieved by model fine-tuning using additional task-specific human demonstration trajectories. The insight of this work is the generalization scenario that the policy is \emph{trained with several shapes, and tested to another set of shapes unseen in the training set, with an additional non-expert cost}. Despite the scenario pushing both the generalization and the robustness against the initial error, the performance of force-based policy is still insufficient in mobile manipulation tasks due to the relatively large base navigation error. 



The vision-based policy is another option, which has the potential to correct the error and make fewer peg-hole contacts. Existing vision-based works \cite{haugaard2020fast} provide mature solutions for holes with 1$mm$ tolerance, which is acceptable for many applications. If sub-$mm$ is required, one can simply employ a follow-up force-based insertion \cite{inoue2017deep}. However, vision-based methods highly depend on manual feature extraction, error modeling, and controller design, struggling to deal with unseen objects. Recent progress in multi-modal representation shows promising results~\cite{lee2019making}, but still faces difficulty to achieve 1$mm$ level precision.

To address the above problem, we propose to bring the generalization to the vision-based policy for holes with 1$mm$ tolerance, from $\sim$10$mm$ initial error. The vision generalization is inspired that the policy is \emph{trained with several shapes, and tested to another set of shapes unseen in the training set, with minimal additional cost, which is a one-minute human pose teaching}. The core idea is to decouple the generalization of the sensory-motor policy to the design of a fast-adaptable perception module and a simulation-based generic policy module. The decoupling is achieved by introducing a segmented image in between. As a result, a framework for the peg-in-hole tasks is proposed, consisting of a segmentation network (SN), a virtual sensor network (VSN), and a controller network (CN). Under this framework, the segmented image-based policy (VSN+CN) is trained totally in simulation, which can perform generic insertion given a peg-hole segmented image, replacing the unsafe and time-consuming real-world training as some force-based policy \cite{inoue2017deep}. The fast-adaption of the perception is achieved by fine-tuning the SN only using hundreds of images. To minimize this additional cost, an automatic data collection and annotation pipeline for image segmentation is designed, which only needs a one-minute human pose teaching. Note that the human effort in this scenario is almost minimal, as it is the same as the traditional position-based robot application. Finally, by connecting the modules, the vision-based policy can be applied to real-world peg-in-hole tasks. In summary, the contributions are as follows: 

\begin{itemize}
\item[1)]
A vision-based peg-in-hole policy framework that decouples the generalization of the sensory-motor policy to the design of a fast-adaptable perception module and a simulation-based generic policy module.
\item[2)]
A segmented-image based insertion policy with VSN and CN, trained totally in simulation, is able to generalize to peg in unseen holes.
\item[3)]
An automatic data collection and annotation pipeline for image segmentation, leverages one-minute human teaching for fast-adaptable perception. 
\item[4)]
Extensive experiments validating the 1$mm$ level performance of the policy, both in the configuration of eye-to-hand and eye-in-hand, as well as a real-world peg insertion application in the EV charging system. 
\end{itemize}

The remainder of this paper is organized as follows: 
Related works are reviewed in Sec.~\ref{rw}. 
The proposed method is detailed in Sec.~\ref{policy}. 
The controller design is presented in Sec.~\ref{controller}. 
Implementations can be found in Sec.~\ref{imp}. Sec.~\ref{exp} provides the simulations and real-world experiments.
Finally, Sec.~\ref{dis} concludes the article.

\section{Related Work}
\label{rw}

\subsection{ Peg-in-hole Assembly}
Robotic peg-in-hole has been studied for several decades from numerous viewpoints, which raises many challenges. 
Conventional line of work like spiral search~\cite{2018Multi} exhaustively locates the hole within an uncertain region following the predefined (scripted) trajectories. However, the long-time contact-rich searching may hurt the surface quality of the peg-hole pairs. 
Model-based insertions~\cite{fei2003assembly} rely on contact model analysis to reason about state conditions and decompose the peg-in-hole tasks into two steps: contact state recognition and compliant control. These algorithms typically either use force/torque signals to detect peg-hole interactions~\cite{Zhang2019Jamming} or leverage vision for pose information~\cite{haugaard2020fast}, based on which optimal insertion trajectories are planned. To overcome the uncertainty in sensing and modeling, compliance control is often applied to the manipulator~\cite{morgan2021vision}. Principally, these work can achieve fast insertion with less contact, but struggle to generalize to new insertion scenarios due to the task-specific design in perception modeling, planning, and control.

Recently with the progress of deep learning, another line of work focus on learning-based methods for the insertion tasks.
\cite{inoue2017deep} achieves high-precision assembly on a tight clearance cylindrical peg-in-hole task by training a reinforcement learning (RL) policy in real-world. The precision required to perform this task exceeds that of robots. But the extensive number of contact-rich interactions increase the possibility to damage the environment and the robot itself, and the learned policy only applies to the cylindrical peg/hole.
To improve sample efficiency and generalization, the human-in-the-loop demonstration has been incorporated as prior information to the RL algorithms~\cite{2021Robust}. However, it takes a long time to transfer the skills to the unseen insertion tasks by online fine-tuning, which hinders the algorithms from quickly scaling up to the real-world unseen insertions.
\cite{lee2019making} shows the potential to generalize over varying geometries by learning a multisensory representation in simulation, based on which an RL-based policy is trained in the real world, which demonstrates to improve the sample efficiency of the policy learning. However, the selected tasks do not require the policy to be as precise as industrial connector insertions. 

\begin{figure*}[]
\includegraphics[height=60mm,width=1\textwidth]{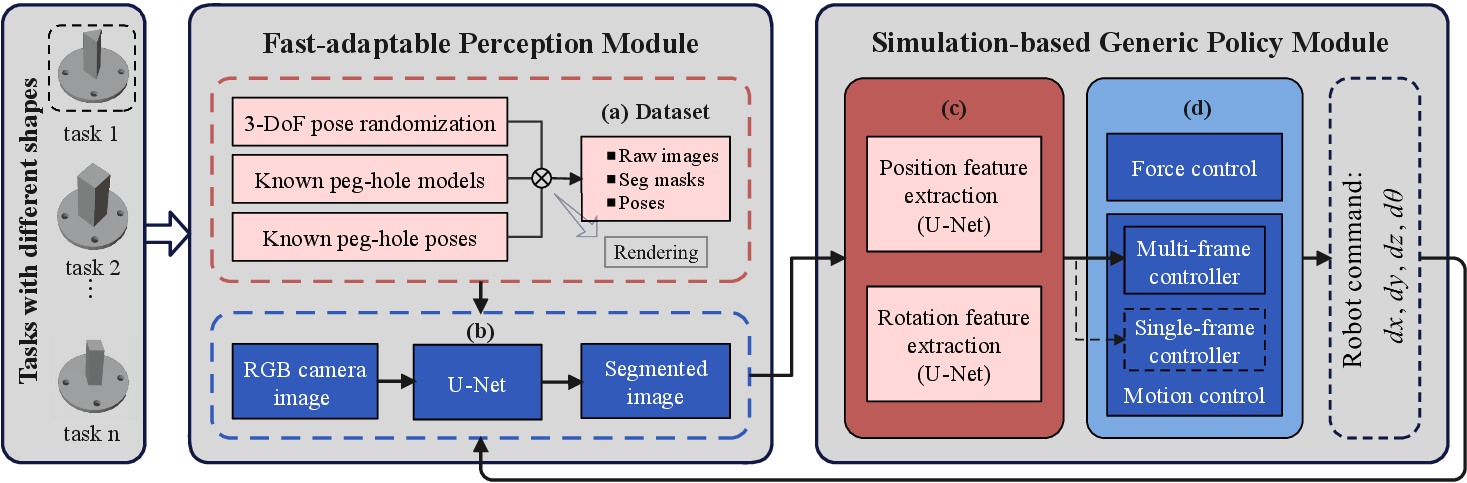}
\caption{The proposed vision feedback framework. 
(a) Automatic data collection and annotation pipeline to accelerate the perception adaptation.
(b) SN: peg-hole segmentation network to transfer the raw RGB image to the segmented image.
(c) VSN: general position and rotation feature extraction based on the segmented image.
(d) The decoupled motion-force control to generate robot command, where the motion control is achieved by the proposed CN.}
\label{framework}
\end{figure*}

\subsection{Data Collection}
The prominent learning-based methodologies employed in contact-rich manipulations, as highlighted in prior research \cite{2010Neural,contact-rich}, specifically in tasks like peg assembly~\cite{primitive2021} and screw fastening~\cite{Puang_2020}, grapple with the well-acknowledged challenge of data scarcity \cite{zhu2018}. In practice, these algorithms often demonstrate optimal performance when supplied with extensive training datasets numbering in the millions, or even billions of instances \cite{2018Reinforcement}.
Regrettably, the current landscape presents a dearth of access to such voluminous datasets. This is largely due to the arduous and resource-intensive nature of data acquisition, characterized by the need for labor-intensive efforts and significant computational resources\cite{2020Meta, 2015Learning2,2018Learning2,2020Variable}. Notably, in the work of Triyonoputro et al. \cite{triyonoputro2019quickly}, a substantial dataset was generated through complex synthetic image generation, which entailed meticulous manual design.
In an effort to mitigate the manual labor involved, researchers have introduced the concept of efficient self-supervised data collection for offline robot learning, employing task-specific loss functions \cite{efficient2021,zakka2020form2fit}. Alternatively, there exists a paradigm of online data collection, achieved without human intervention, where robots interact autonomously with their environments \cite{2014Reinforcement}. Nevertheless, despite these advancements, it remains a time-consuming endeavor to accumulate the vast quantities of real-world data required for optimal performance.
Furthermore, it is imperative to acknowledge the safety concerns inherent in the wide-ranging random sampling often associated with trial and error, as it poses risks not only to the physical integrity of the robot but also to the integrity of the contact environment itself \cite{7576815}.

\subsection{Vision Feedback Manipulation}
With recent advances in computer vision~\cite{survey}, a lot of prior work focuses on learning an end-to-end visuomotor controller that directly maps raw pixel observations to control actions and closes the control loop with vision feedback~\cite{dong2021tactile}. They have shown impressive success in exploring high-dimensional environments to learn complex robot manipulation tasks. However, these end-to-end algorithms require a large amount of training data and remain difficult to generalize to new scenarios without extensive fine-tuning. 
Borrowing the idea from another line of research that decouples the system into individual sensing and control policy~\cite{lee2019making}, our system is separated into a visual perception module and a generic policy module by a segmentation network in between, based on which the policy generalization problem is transferred to the perception adaptation. While many of these methods introduce the decoupling mechanism with the aim to better incorporate the state-of-the-art vision-based techniques~\cite{morgan2021vision, stevvsic2020learning, haugaard2020fast} (e.g. pose estimation and tracking, etc.), we instead leverage it as a way to achieve the two sub-tasks: simulated generic policy and fast sim-to-real.

Some vision-based algorithms are limited to the eye-to-hand configuration in order to obtain the robust depth information~\cite{zakka2020form2fit}, and many other methods require the camera to move with the robot arm for active sensing~\cite{vision2022}, which only fits the eye-in-hand configuration. Our method, in contrary, applies to broader application scenarios where the camera can be mounted under either the eye-to-hand or eye-in-hand configurations. 



 \begin{figure}[t]
\centering
\includegraphics[height=87mm,width=0.48\textwidth]{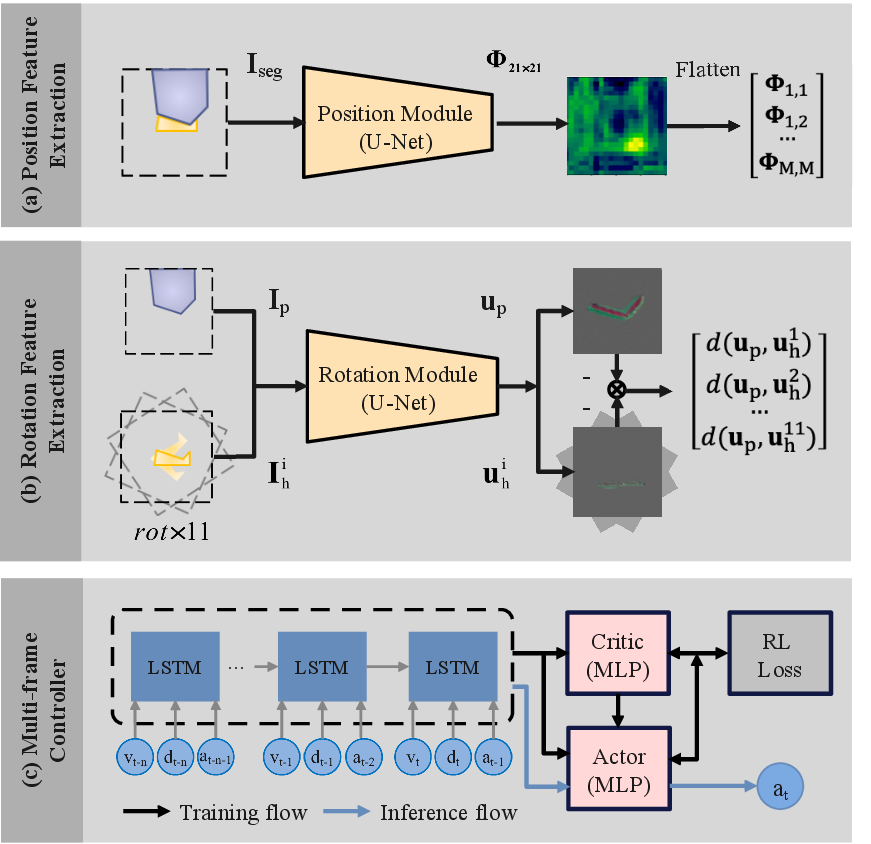}
\caption{(a) The VSN for position feature extraction. (b) The VSN for rotation feature extraction. (c) The CN for insertion trajectory planning. All modules are learned in simulation.}
\label{network}
\end{figure}

\subsection{Transfer Learning}
Transfer learning is an optimization that allows rapid progress or improved performance when modeling the second task\cite{Pan2010ASO,yosinski2014transferable}. In the field of computer vision, models are firstly pretrained in large dataset such as the ImageNet\cite{5206848}, and then transferred to other tasks by retraining partial layers of the model in the target dataset to accelerate the training process, seeing that features are assumed to be more general in early layers and more original-dataset-specific in later layers\cite{pmlr-v27-bengio12a}. For natural language processing tasks, efficient algorithms exist to pretrain a word embedding on very large corpa of text documents such as Google’s word2vec\cite{mikolov2013efficient} model and GloVe Model from Stanford\cite{pennington2014glove}, and use them for downstream tasks with few-shot learning\cite{du2021fewshot}. 

Sim-to-real is the branch of the transfer learning, which focus on the sample efficiency and safety in the field of robotics when the robot interact with the environment 
in real world. 
Simulation is leveraged to improve sample efficiency and safety in the data collection process~\cite{2018Learning, 2017Sim, 2018Using} but also inducts the challenge of the reality gap. 
Recently, successful sim-to-real transfer for perception\cite{danielczuk2019segmenting}, grasping\cite{mahler2018dexnet}, and feedback control policies\cite{openai2019solving} includes: domain adaption\cite{Tzeng2015TowardsAD}, domain randomization\cite{haugaard2020fast,tobin2017domain}, and progressive network\cite{rusu2018simtoreal}. Despite progress, a lot of care has to be taken manually to make the simulation as realistic as possible and choose the parameters to randomize over.
\cite{ding2019transferable} propose an inverse dynamic model which is learned with the force/torque data generated in a sample-efﬁcient offline fashion. To achieve sim-to-real, the dynamic model is fine-tuned and transferred to new peg/holes after a small number of insertion trials.
Borrowing the ideas from transfer learning, our system achieves fast sim-to-real by decoupling the visual perception and the policy control. The policy control gains the ability to be general to various insertion tasks by training in simulation with extensive data. And the visual perception enjoys few-shot retraining to adapt to specific downstream tasks in real world with a much smaller amount of data (500 image frames). Under this framework, we achieve sim-to-real efficiently without manual effort.

\section{Methods}
\label{policy}
The proposed framework integrates the following modules to complete real-world insertion tasks for peg in unseen holes: a fast-adaptable perception module and a simulation-based generic policy module. As depicted in Fig.~\ref{framework}, the generic policy consists of a VSN and a CN for insertion tasks, which are trained solely in simulation. 
Then given an unseen real-world insertion task, 
only the SN has to be fine-tuned, to transfer the RGB image to the segmented image and provide feedback for the control loop. 
To further accelerate the adaptation, an automatic data collection and annotation pipeline for the SN is proposed.



\subsection{Virtual Sensor Network}
The VSN functions as a virtual sensor that tries to extract the position and rotation features respectively for pose alignment. Loss functions are delicately designed with inductive bias introduced to encourage the VSN to learn the related features. We represent the features based on spatial heatmaps, which allow for better spatial generalization than directly regressing point coordinates or angles from a latent representation as shown in \cite{2018Numerical}.

\subsubsection{Position Feature Extraction}
Uses a deep neural network that takes the segmented image $\mathbf{I}_{\rm seg}$ as input. $\mathbf{I}_{\rm seg}$ only contains the pixels of the peg, the hole
and, the background, which can be obtained from simulation directly. The spatial heatmaps are generated to predict the peg-centric (eye-in-hand) or the hole-centric (eye-to-hand) 2D planar displacement. As shown in Fig. \ref{network}(a), the position of the brightest point indicates the relative pose between the peg and the hole. We choose U-Net~\cite{unet} $f_\varphi$ parameterized by $\varphi$ as the backbone to predict the 2-channel one-hot heatmaps from $\mathbf{I}_{\rm seg}$ in Eq. \ref{seg_eq}:
\begin{equation}
\mathbf{\Phi}=f_{\varphi}(\mathbf{I}_{\rm seg}) \label{seg_eq}
\end{equation}

Let $dx^*$ and $dy^*$ be the ground truth of pixel planar displacement along the X-axis and Y-axis respectively, which can be calculated by projecting from the 3D world coordinates with the calibrated camera intrinsic and extrinsic matrices. Then the desired pixel value $\mathbf{\Phi}_{\rm i,j}^*$ can be defined in Eq. \ref{gt},  where $i,j$ are the pixel coordinates of the ground truth heatmaps with the image center to be the coordinate origin.
\begin{equation}
\mathbf{\Phi}_{\rm i,j}^* = 
\begin{cases}
1 & \text{if $i = dx^*$, $j= dy^*$}  \\ 
0 & \textit{o.w.}
\end{cases}
\label{gt}
\end{equation}

The loss function is defined over the ground truth heatmaps $\mathbf{\Phi}^*$ and the predicted heatmaps $\Phi$ with binary cross entropy loss as in Eq. \ref{pos-loss}.
\begin{gather}
Loss_{\varphi}=-\sum(\mathbf{\Phi}^*\times\log\mathbf{\Phi} + (1-\mathbf{\Phi}^*)\times\log(1-\mathbf{\Phi})) \label{pos-loss}
\end{gather}

The heatmap can be leveraged not only as the value function as in RL, where we can extract the policy (defined as the VSN policy) for position alignment, but also as the features with uncertainty for the CN (detailed in \ref{planner}). For the inference of the VSN policy, the point with the largest value in the produced heatmap is selected as the predicted point, whose indexes are the corresponding translate $dx, dy$ as:
\begin{equation}
dx, dy = \mathop{\arg\max}\limits_{i,j}\mathbf{\Phi}_{\rm i,j}\label{pos}
\end{equation}
To obtain the features for the CN, the 2D heatmap is flattened to the 1D vector as defined in Eq. \ref{value}.
\begin{equation}
\label{value}
\mathbf{v} = \left\{\mathbf{\Phi}_{\rm i,j}|i,j\in \mathbb{N}, i,j\in[0,M) \right\}
\end{equation}
where $\mathbf{\Phi}$ is the pixel value of the heatmap, and $M$ indicates the spatial size of the heatmap.

\subsubsection{Rotation Feature Extraction}
Uses the U-Net $f_{\psi}$ as the backbone which consists of a two-stream Siamese network~\cite{ROOPAK1993SIGNATURE} with shared parameters $\psi$ (see Fig. \ref{network}(b)). The first stream takes as input the segmented image $\mathbf{I}_{\rm p}$ which only contains the pixels of the peg and the background. The second stream takes as input the 11 rotated segmented images $\mathbf{I}_{\rm h}^{\rm i}$ which only contain the pixels of the hole and the background, with 5 equispaced clockwise rotations and 5 equispaced anticlockwise rotations. Both streams output feature maps with the same resolution as the input, defined as $\mathbf{u}_{\rm p}, \mathbf{u}_{\rm h}$ respectively. The $11\times1$ feature vector produced by the rotation module is defined as:
\begin{gather}
    \mathbf{u}_{\rm p} = f_{\psi}(\mathbf{I}_{\rm p})\\
    \mathbf{u}_{\rm h}^{\rm i} = f_{\psi}(\mathbf{I}_{\rm h}^{\rm i})\\
    \mathbf{d} = \left\{e^{-d(\mathbf{u}_{\rm p}, \mathbf{u}_{\rm h}^{\rm i})}|i\in \mathbb{N}, i\in [0,11)\right\}\label{dis}
\end{gather}
where $d$ is the Euclidean distance between the two feature maps. Among the 11 rotated images $\mathbf{I}_{\rm h}^{\rm i}$, only the correct one $\mathbf{I}^{\rm p}_{\rm h}$ with the ground-truth rotation will match $\mathbf{I}_{\rm p}$ and the others $\mathbf{I}^{\rm n}_{\rm h}$ will mismatch. To establish the rotation correspondences between the peg and the hole, a variant triplet loss function~\cite{BMVC2016_119} is designed to encourage the paired feature maps to match, while pushing the unpaired maps apart. By constraining the feature distance (defined in Eq. \ref{dis}) in the range of $(0,1]$, the paired feature distance $\mathbf{d}_{\rm p}$ is expected to be as close to 1, while the unpaired distance $\mathbf{d}_{\rm u}$ to be as close to 0.

\begin{equation}
    Loss_{\psi} = \max\{\mathbf{d}_{\rm u}-\mathbf{d}_{\rm p}+1, 0\}
\end{equation} 

As with the position module, the 1D feature vector can be leveraged not only as the VSN policy for rotation alignment but also as the features for the CN. For the inference of the VSN policy, the predicted rotation $d\theta$ can be obtained by finding the maximum distance from $\mathbf{d}$ as:
\begin{equation}
    d\theta = \beta\times(\mathop{\arg\max}\limits_i\mathbf{d}_{\rm i}-5)
    \label{rot}
\end{equation}
where $\beta$ is the rotation resolution. Alternatively, the feature vector $\mathbf{d}$ can be taken as input by the CN.

\subsection{Controller Network}
\label{planner}
We design the CN architecture with reinforcement learning to plan a feasible and efficient insertion trajectory based on the VSN features. We model the insertion task as a finite-horizon, discounted Markov decision process $\mathcal{M}$, with a state space $\mathcal{S}$, an action space $\mathcal{A}$, a reward function $r: \mathcal{S}\times{\mathcal{A}} \to \mathbb{R}$ and a discount factor $\gamma\in (0,1]$. We use the policy gradient-based algorithm as it is more stable to learn. The RL agent starts from trial and error to generate a trajectory with random actions $a$. By increasing exploitation and reducing exploration over time, the agent strives to maximize the expected cumulative rewards:
\begin{equation}
    R(\tau) = \sum_{t=0}^{T-1}\gamma^tr(\mathbf{s}_{\rm t},\mathbf{a}_{\rm t})
\end{equation}
\begin{equation}
    J(\pi_{\theta})=\mathbb{E}_{\tau\sim\pi_{\theta}}\left[R(\tau)\right]\label{eq1}
\end{equation}
where $\tau$ is the trajectory, $T$ is the trajectory length, $\pi_{\theta}$ is the policy parameterized by $\theta$, $\gamma$ is the defined discount factor, $r$ is the defined reward function, $\mathbf{s}_{\rm t}$ is the defined system state, and $\mathbf{a}_{\rm t}$ is the defined action. By calculating the derivation of (\ref{eq1}), we define the loss function for RL as:
\begin{equation}
    \nabla_{\pi_{\theta}}J(\pi_{\theta})=\mathbb{E}_{\tau\sim\pi_{\theta}}\left[\sum_{t=0}^{T}\nabla_{\pi_{\theta}}\log\pi_{\theta}(\mathbf{a}_{\rm t}|\mathbf{s}_{\rm t})R(\tau)]\right]
\end{equation}
To train the RL agent, we collect the trajectory dataset by running the policy online, computing the gradient with backpropagation, and performing policy updates. We propose two types of controllers, the single-frame controller, and the multi-frame controller. Both of the two controllers are based on the Advantage Actor-Critic \cite{mnih2016asynchronous}, which is a modern RL algorithm based on policy gradient.

\subsubsection{Single-frame Controller}
Takes as input the concatenated features of the position and the rotation modules defined as:
\begin{equation}
    \mathbf{s}_{\rm t} = [\mathbf{v}, \mathbf{d}]\label{eq6}
\end{equation}
The controller $\pi$ generates an action $\mathbf{a}_{\rm t}$ as the robot command:
\begin{equation}
    \mathbf{a}_{\rm t} = [a_x, a_y, a_{\theta}]_t =\pi(\mathbf{s}_{\rm t})
\end{equation}
A dense reward $r$ is obtained from the environment after performing each step of the robot command:
\begin{equation}
r = 
\begin{cases}
1  & success  \\ 
-\frac{1}{k_{max}} & \textit{o.w.}
\end{cases}
\label{reward}
\end{equation}
where $k_{max}$ is the preset maximum step of an trajectory. The reward is designed to encourage the policy to complete the insertion as fast as possible. The trajectory finishes when total steps reaching $k_{max}$ or successful insertion is achieved. 

\subsubsection{Multi-frame Controller}
In the single-frame controller, feature occlusion often occurs which hinders the policy to make accurate estimations. Thus, the state is not fully observable from a single time instant. We propose the multi-frame controller to reduce the one-frame perception uncertainty by aggregating the temporal features and actions over time (see Fig. \ref{network}(c)). We introduce the LSTM network $\mathcal{L}$ parameterized by $\upsilon$ to encode the historical sequential features and actions. By concatenating the state of the single-frame controller with the action and stacking them with the latest n-steps of features and actions, the current state is redefined as: 
\begin{equation}
    \mathbf{s}_{\rm t} = \mathcal{L_{\upsilon}}([\mathbf{v}_{\rm t-n:t}, \mathbf{d}_{\rm t-n:t}, \mathbf{a}_{\rm t-n-1:t-1}])
\end{equation}
The state is then taken as input for the RL agent to generate the current action $\mathbf{a}_{\rm t}$.

\begin{figure}[]
\centering
\includegraphics[height=105mm,width=0.48\textwidth]{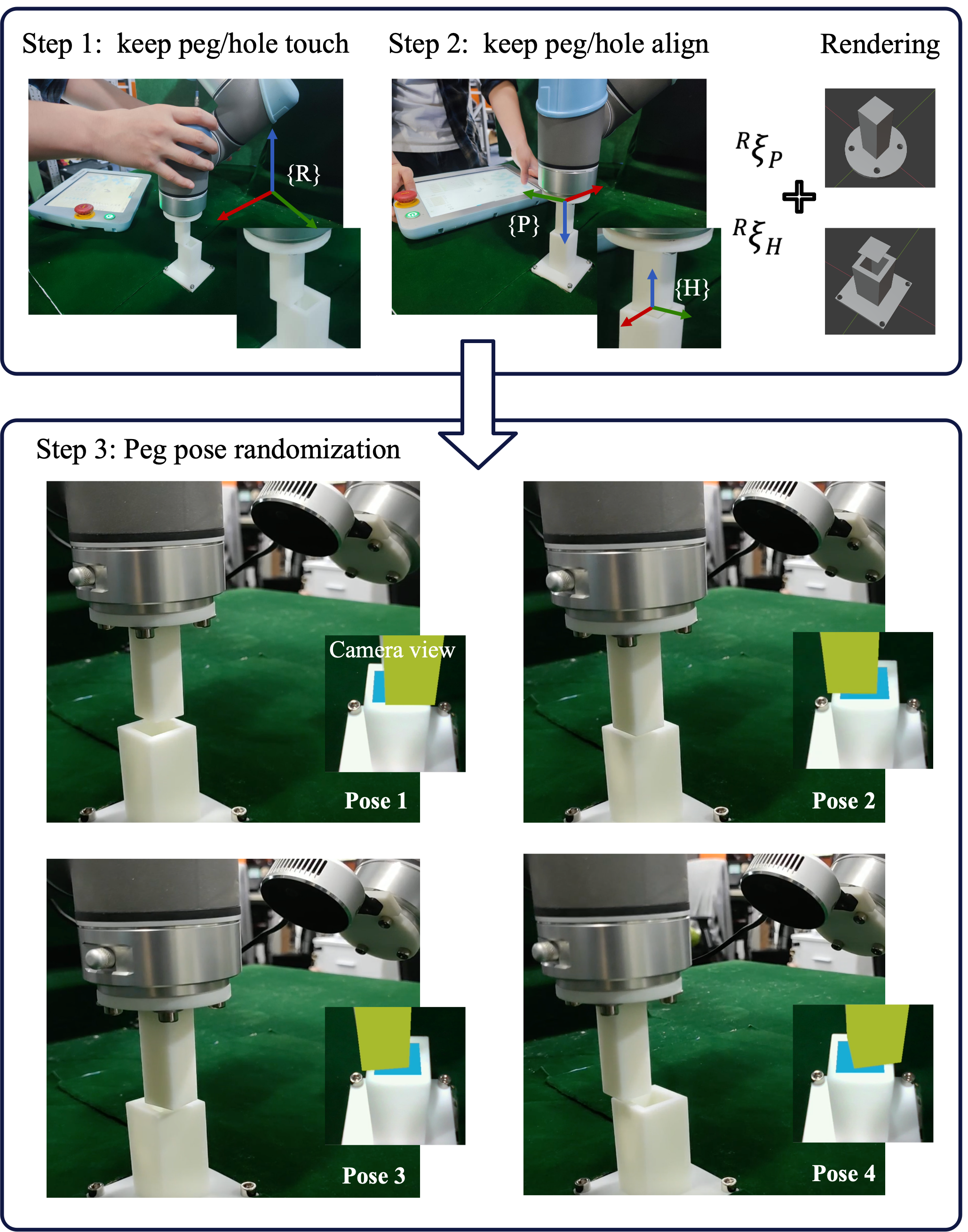}
\caption{Automatic data collection and annotation pipeline with one-minute non-expert teaching.}
\label{automatic}
\end{figure}

\subsection{Segmentation Network}
\subsubsection{Peg-hole Segmentation}
We use the U-Net $f_{\phi}$ parameterized with $\phi$ as the backbone of the SN. The SN takes as input the raw RGB image $\mathbf{I}_{\rm rgb}$ from a real-world calibrated camera, and outputs the 3-channel prediction $\mathbf{I}_{\rm out}$ with the same resolution of the input as:
\begin{equation}
    \mathbf{I}_{\rm out} = f_{\phi}(\mathbf{I}_{\rm rgb})
\end{equation}
The loss function is defined over the ground truth $\mathbf{I}^*$ and the prediction $\mathbf{I}_{\rm out}$ with cross entropy loss as in:
\begin{equation}
Loss_{\phi}=-\sum \mathbf{I}^*\times\log\mathbf{I}_{\rm out}
\end{equation}
The 1-channel segmented image $\mathbf{I}_{\rm seg}$ can be obtained from $\mathbf{I}_{\rm out}$ by choosing the index with the max value along the channels as:
\begin{equation}
    \mathbf{I}_{\rm seg}^{\rm i,j}= \mathop{\arg\max}\limits_{k}\mathbf{I}_{\rm out}^{\rm i,j,k}
\end{equation}
where $i,j$ are the pixel coordinates of the segmented image.

\begin{figure}
\centering
\includegraphics[height=52mm,width=0.48\textwidth]{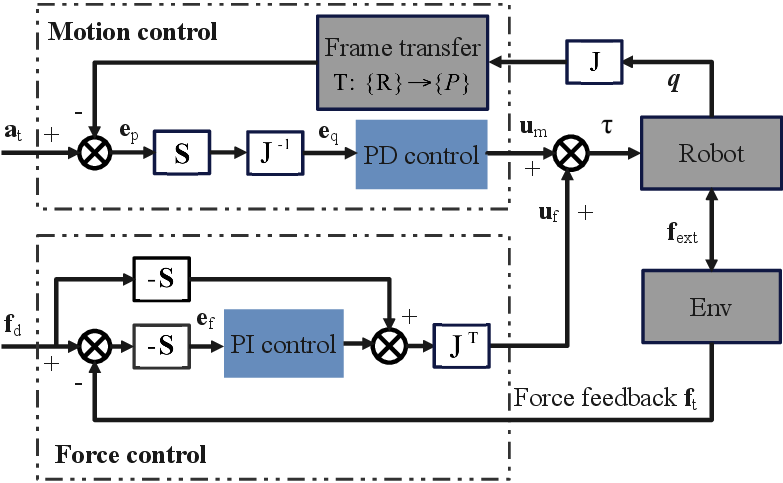}
\caption{Decoupled motion-force controller}
\label{control}
\end{figure}

\begin{table}
\centering
\caption{Task Coefficients}
\begin{tabular}{ll}
\hline
\multicolumn{1}{c}{Coefficient} & \multicolumn{1}{c}{Value} \\ \hline
Position heatmap size ($M$) & 21 pixels \\
Rotation resolution ($\beta$) & 2$^{\circ}$ \\
Number of LSTM layers & 1 \\
Hidden size of LSTM layer & 32 \\
Number of Historical frames ($n$) & 5 \\
Maximum trajectory length ($k_{max}$) & 100 \\
Discount factor ($\gamma$) & 0.995 \\
PD proportional gain ($k_{pm}$) & 20 \\
PD differential gain ($k_{vm}$) & 5 \\
PI proportional gain ($k_{pf}$) & 0.004 \\
PI integral gain ($k_{if}$) & 1 \\
Constant contact force ($\mathbf{f}_{\rm d}$) & 10N \\ \hline
\end{tabular}
\label{coeff}
\end{table}

\subsubsection{Data Collection and Annotation}
A crucial step of the algorithm presented in the previous section is collecting sufficient real-world training data for the perception adaptation, which would be impractical given the fact that the time-consuming, labor-intensive manual annotation for these data can cost a few days. In this section, we design an automatic data collection and ground truth annotation pipeline for the SN fine-tuning, which helps accelerate the adaption to various real-world insertion tasks.  

We define the world frame as $\left\{R\right\}$ by taking the robot base as the origin in the coordinate system. We also define the reference frame as $\left\{P\right\}$ which is attached to the robot end-effector, and the reference frame $\left\{H\right\}$ which indicates the hole pose. 
The pipeline has three steps as illustrated in Fig.~\ref{automatic}. 
Firstly, the peg is kept in touch with the hole by moving the robot end-effector with a human demonstration. 
Secondly, by controlling the direction key of the teaching pendant, the peg-hole pose can be aligned precisely. After completing the two steps, the peg/hole poses in the robot coordinate system can be obtained. Concretely, The peg can be regarded as the robot end-effector in the robot-peg system, where the peg pose $^R\xi_{P}$ can be controlled and calculated by robot kinematics. The hole is fixed in the world space, and the hole pose $^R\xi_{H}$ can be obtained by robot kinematics with the current joint angles. With the known 3D peg/hole models and known peg/hole poses, the peg/hole masks can be obtained by rendering the 3D models with their poses in Pyrender~\cite{pyrender}.
Finally, by controlling and randomizing the peg pose relative to the hole, a training dataset consisting of the RGB images, their paired masks, and the corresponding relative poses can be obtained.
Notably, the masks are annotated for training the SN and the relative poses can be leveraged for fine-tuning the VSN.
The whole process only needs one-minute human teaching to get the hole pose with all the data collection and annotation done by the robot automatically, which is time-efficient and manually cost-minimal.

\begin{figure*}
\centering
\includegraphics[width=0.99\textwidth]{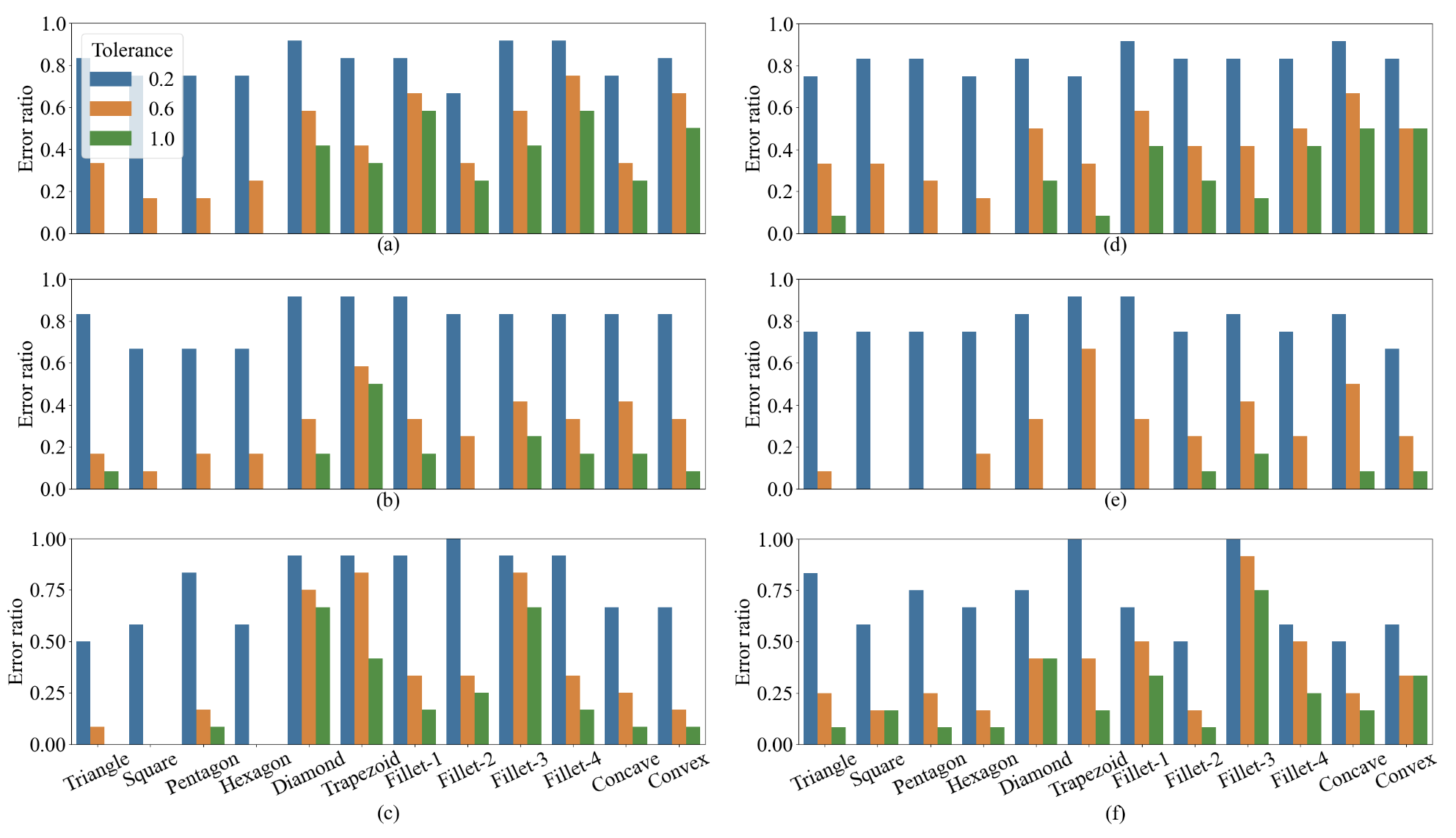}
\caption{VSN prediction error under increasing tolerances (0.2$mm$, 0.6$mm$, 1.0$mm$)  on (a) $x$-axis with the eye-to-hand configuration (b) $y$-axis with the eye-to-hand configuration (c) $yaw$ rotation with the eye-to-hand configuration (d) $x$-axis with the eye-in-hand configuration (e) $y$-axis with the eye-in-hand configuration (f) $yaw$ rotation with the eye-in-hand configuration.}
\label{sensor_precision}
\end{figure*}

\section{Controller Design}
\label{controller}
The reference frame axes in {P} indicates the 6-DoF peg pose, which consists of the end-effector position $\textbf{x}\in \mathbb{R}^{3}$ and the end-effector rotation $\mathbf{R}\in \mathbf{SO}(3)$. The produced actions of the policy are represented in the reference frame $\left\{P\right\}$. Our policy solves the 4-DoF peg-in-hole which consists of the 3-DoF position $\textbf{x}$ and the $z$-axis yaw rotation $\theta$.

We use the hybrid motion-force controller to execute the end-effector control commands (see Fig. \ref{control}). For the motion control, the desired action $\mathbf{a}_{\rm t}$ produced by the policy is executed by a PD controller with the control law defined as:
\begin{gather}
    \mathbf{e}_{\rm q} = \mathbf{S}\mathbf{J}^{\rm -1}(\mathbf{a}_{\rm t} - \mathbf{T}_R^{P}\mathbf{Jq})\\
    \mathbf{u}_{\rm m} = k_{pm}\mathbf{e}_{\rm q} + k_{vm}\mathbf{\dot{e}}_{\rm q}
\end{gather}
where $\textbf{S}$ is the diagonal matrix to decouple the motion and force control, $\textbf{J}$ is end-effector Jacobian, $\textbf{T}$ is the transform matrix from $\left\{R\right\}$ to $\left\{P\right\}$, $\textbf{q}$ is joint displacement in joint space, $k_{pm}$ and $k_{vm}$ are proportional gain and differential gain respectively. For the force control, constant contact force $\mathbf{f}_{\rm d}$ between the peg and the hole is expected to be maintained by a PI controller with the control law defined as:
\begin{gather}
    \mathbf{e}_{\rm f} = -\mathbf{S}(\mathbf{f}_{\rm d}-\mathbf{f}_{\rm t})\\
    \mathbf{u}_{\rm f} = \mathbf{J}^{\rm T}(k_{pf}\mathbf{e}_{\rm f}+k_{if}\int\mathbf{e}_{\rm f}-\mathbf{{S}}\mathbf{f}_{\rm d})
\end{gather}
where $\mathbf{f}_{\rm t}$ is the force feedback from the F/T sensor, $k_{pf}$ and $k_{if}$ are the proportional gain and the integral gain respectively.





\section{Implementations}
\label{imp}

We build a simulation environment in PyBullet \cite{erwin2019Python} for model training. We use the Franka Panda robot, a 7-DoF torque-controlled robot, of which the end-effector is connected by a peg with a fixed joint. A wrist F/T sensor is mounted to measure the contact force. An RGB camera is either fixed at the world space following \cite{haugaard2020fast} (eye-to-hand) or mounted at the robot end-effector\cite{triyonoputro2019quickly} (eye-in-hand). 
In addition, we also evaluate the proposed method on a real-world platform, which consists of a 6-DoF UR5 robot, a Robotiq FT300 wrist F/T sensor, and a calibrated camera of Intel RealSense L515. We use Pyrender, a physically-based Python library, to obtain the segmentation masks by rendering the 3D models. 

We use a single Nvidia GeForce GPU (RTX 2070) with an Intel Core (i7 8700) for data collection and model training.
For each insertion trajectory, early termination occurs on successful insertion or if the robot end-effector leaves the safe region. We set 50$mm$ for the maximum safe distance.
The RGB image resolution is cropped to 224$\times$224. 

The initial position error ranges from -10$mm$ to 10$mm$ and the initial rotation error ranges from $-10^\circ$ to $10^\circ$. To evaluate the shape generalization, 4 seen peg shapes (Fig. \ref{setting}(b)) are designed for training and 8 unseen peg shapes (Fig. \ref{setting}(c)) for evaluation. The 8 unseen peg shapes are selected from general daily insert tasks which range from simple to complex, concave to convex, and square-corner to round-corner. More details can be found in Tab.~\ref{coeff}.

\section{Experiments}
\label{exp}
We design a series of experiments to examine the effectiveness and practicability of the proposed framework. In particular, our goal is to evaluate the generalization ability on the unseen holes and the fast adaptation attributes. Experiments are designed and conducted to answer the following five questions: 

\begin{itemize}
\item [1)]
Generalization: Is the policy able to generalize to the objects that are unseen in the training set?
\item [2)]
Efficiency: How fast the vision feedback peg-in-hole can be under 10$mm$ initial error on the unseen shapes?
\item [3)]
Precision: What is the performance trend of the sub-$mm$ peg-in-hole with decreasing tolerances?
\item [4)]
Robustness: Is the policy able to plan a feasible trajectory under vision occlusion with only one camera?
\item [5)]
Adaptation: How fast the proposed system can be adapted from simulation to different unseen real-world scenarios?
\end{itemize}

\begin{figure}[t]
\centering
\includegraphics[height=82mm,width=0.48\textwidth]{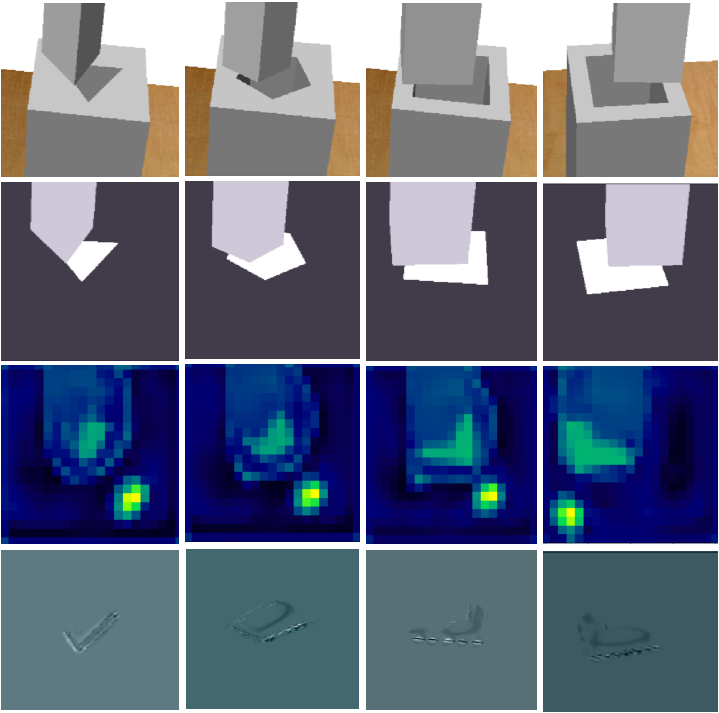}
\caption{Feature visualization of the VSN: raw images(top), segmented images(upper-middle), position feature maps (lower-middle) and rotation feature maps (bottom).}
\label{feature}
\end{figure}

\subsection{Simulations}

\subsubsection{Generalization of the VSN}
\label{snn_precision}
We evaluate the generalization of the VSN by calculating the VSN precision on the seen and unseen holes respectively. The VSN precision is evaluated from three dimensions: $x,y$-axis and $yaw$ rotation respectively, by comparing the predicted value $dx, dy$ and the rotation $d\theta$ with the corresponding ground truth value $v^*$ as:

\begin{gather}
\label{precision}
    Error~ratio~(v) = \frac{\#\{trials~with~ |v-v^*|>tol\}}{\#\{total~trials\}} \\
    v=dx,~dy,~or~d\theta 
\end{gather}
where $\#\{\cdot\}$ means the trial number and $tol$ is the hole tolerance.
We evaluate the VSN precision with three different hole tolerances: 0.2$mm$, 0.6$mm$ and 1.0$mm$.
The results are the average of 12 random trials for each shape under both the eye-to-hand and the eye-in-hand configurations. 
As shown in Fig. \ref{sensor_precision}, 
we observe that the VSN has a high precision on the $y$-axis with the averaged error ratio under 0.2 with 1.0$mm$ tolerance.
The VSN performs relatively poorly on the $x$-axis, where the camera is mounted, due to the reason that the prediction on the $x$-axis relies heavily on the accurate mapping from the pixel distance to the real world displacement. However, the VSN still reduces the averaged error ratio to 0.4 with 1.0$mm$ tolerance.
We also observe that for the $yaw$ rotation under the eye-to-hand configuration, the peg outline changes as the robot moves, which is harder for the VSN to extract the line/corner features for the rotation alignment, but the VSN achieves a high precision with the error ratio lower than 0.25 for most shapes with 1.0$mm$ tolerance.
It is notable that with sub-$mm$ tolerance~(0.6$mm$), which is much more challenge than the 1.0$mm$ tolerance setting, the VSN achieves competitive performances on all the three dimensions, particularly under the seen shapes.

\subsubsection{Feature Visualization of the VSN}
To figure out what has been learned inside the VSN, we perform feature visualization to display the outputs of the VSN. Fig. \ref{feature} visualizes the results, where the top row contains the raw images captured from the RGB camera under four different conditions, where the first three columns are three different peg/hole pairs with the same initial pose, and the last two columns share the same peg/hole shape but are initialized with different poses. The upper-middle row is the corresponding segmented images, the lower-middle row are the one-hot heatmaps produced by the position feature extraction module, and the bottom row is the feature maps of the rotation feature extraction module. 
For the position feature extraction, we can find that the position of the brightest point in the heatmap keeps track of the hole,
which indicates that the position module focuses on the relative position between the peg-hole while avoiding over-fitting to the geometry features. For the rotation feature extraction, the feature maps have learned to encode the general features (corner, line features) of the segmented image for rotation alignment. In general, the designed VSN architecture is insensitive to the shape differences but learns to encode general features to determine the relative position and rotation between the peg and the hole, which explains the shape generalization of the proposed VSN.


\subsubsection{Peg-in-hole with 0.1$\rm\sim$1\textit{mm} Tolerance}
\label{submm}
We compare the proposed policy (VSN+CN) with the following two alternative methods: the VSN (alone) and the E2E Vision-RL~\cite{schoettler2020deep}. 
The same 4 seen peg/hole pairs are selected to train the policy and 8 unseen pairs are selected for evaluation.

\textbf{VSN Policy}:
Since the VSN has explicitly produced the translation $dx,dy$ in Eq. \ref{pos} and the rotation $d\theta$ in Eq. \ref{rot}, we can perform one-shot insertion by moving the robot-peg to the target pose calculated by $dx, dy, d\theta$. 

\textbf{E2E Vision-RL}:
Takes the segmented image instead of the raw RGB image as input as with our proposed method for a fair comparison. The policy network, action space, and reward function is defined the same as in \cite{schoettler2020deep}.

\begin{figure}[t]
\centering
\includegraphics[width=0.48\textwidth]{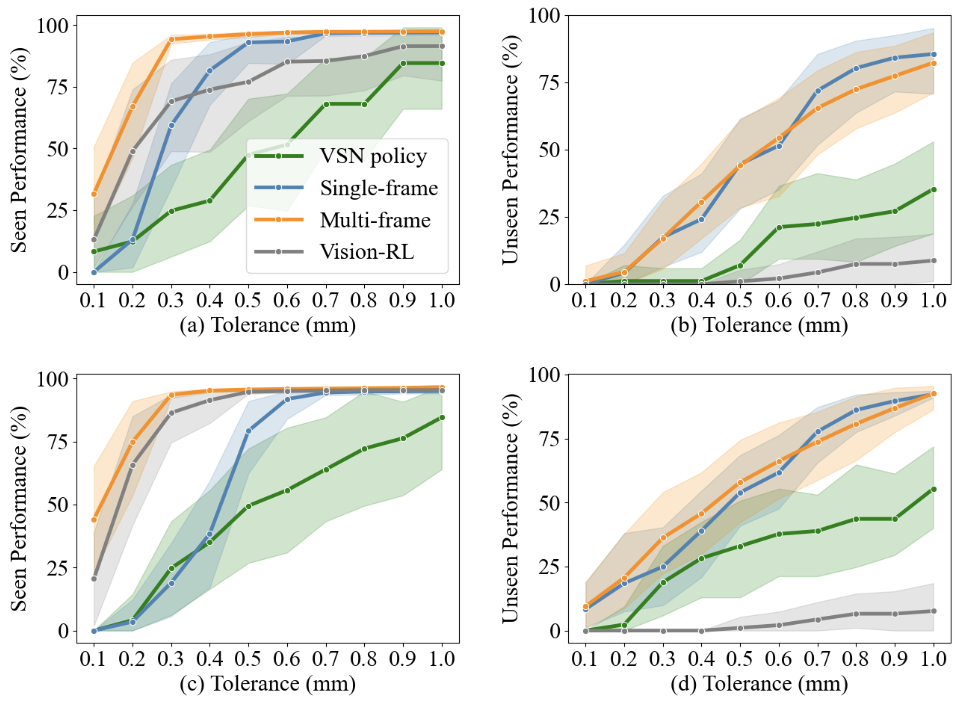}
\caption{Sub-$mm$  Insertion performance  averaged on (a) seen shapes under eye-to-hand (b) unseen shapes under eye-to-hand (c) seen shapes under\textbf{} eye-in-hand (d) unseen shapes under eye-in-hand configuration.}
\label{success_rate}
\end{figure}

We explore the insertion performance with sub-$mm$ tolerance. The performance is evaluated by counting the total rewards (defined in Eq. \ref{reward}) of the full insertion trajectory. The reward is designed to encourage the policy to achieve success insertion with minimal steps. Thus the performance reflects both the success rate and insertion efficiency. 
The hole tolerance ranges from 0.1 to 1$mm$ with 10 steps. Results are averaged over 4 seen and 8 unseen shapes respectively under both the eye-to-hand and eye-in-hand configurations, with each shape performing 12 random insertion trials, which counts to 2880 trials in total. 
As shown in Fig. \ref{success_rate}, the performance curve ascends with the increasing tolerance on both the seen and the unseen shapes and under both the eye-in-hand and the eye-to-hand configurations. 
However, we find that the E2E Vision-RL performs well on the insertions with the seen shape, while struggling to generalize to the unseen shapes.
Compared with the VSN policy, our proposed policies (the single-frame and the multi-frame policy) help promote the insertion performance, though they rely on the VSN to generalize to the unseen shapes. 
In addition, we find that the performance of the multi-frame policy exceeds that of the single-frame policy on the seen shapes, while their performances are level pegging on the unseen shapes. We assume that the superiority of the multi-frame policy over the single-frame policy is based on the higher precision of the VSN. 

\begin{table*}[t]
\centering
\setlength{\tabcolsep}{3.9pt}
\caption{Insertion Performance with 1$mm$ Tolerance in Simulation}
\begin{threeparttable}[t]
\begin{tabular}{ccccccccccccccc}
\hline
 & \multicolumn{5}{c}{Seen} & \multicolumn{9}{c}{Unseen shape} \\
\multirow{-2}{*}{Methods} & 1 & 2 & 3 & 4 & Average & 1 & 2 & 3 & 4 & 5 & 6 & 7 & 8 & Average \\ \hline
\textbf{} & \multicolumn{14}{c}{Success rate} \\ \hline
VSN policy & 10/12 & 12/12 & 9/12 & 10/12 & 85.4\% & 3/12 & 1/12 & 4/12 & 6/12 & 2/12 & 4/12 & 8/12 & 6/12 & 35.4\% \\
VSN policy$^{a}$ & 9/12 & 11/12 & 11/12 & 10/12 & 85.4\% & 5/12 & 5/12 & 8/12 & 8/12 & 4/12 & 8/12 & 8/12 & 9/12 & 57.3\% \\
E2E vision-RL & 12/12 & 12/12 & 12/12 & 12/12 & 100\% & 2/12 & 3/12 & $\times$ & $\times$ & $\times$ & $\times$ & $\times$ & 2/12 & 7.3\% \\
E2E vision-RL$^{a}$ & 11/12 & 12/12 & 12/12 & 11/12 & 95.8\% & 3/12 & 5/12 & $\times$ & $\times$ & $\times$ & $\times$ & $\times$ & $\times$ & 8.3\% \\
\textbf{Single-frame} & 12/12 & 12/12 & 12/12 & 12/12 & {\color[HTML]{000000} \textbf{100\%}} & 12/12 & 12/12 & 12/12 & 12/12 & 12/12 & 11/12 & 12/12 & 4/12 & {\color[HTML]{000000} \textbf{92.7\%}} \\
\textbf{Single-frame$^{a}$} & 12/12 & 12/12 & 12/12 & 12/12 & {\color[HTML]{000000} \textbf{100\%}} & 12/12 & 12/12 & 12/12 & 12/12 & 12/12 & 11/12 & 9/12 & 8/12 & {\color[HTML]{000000} \textbf{93.8\%}} \\
\textbf{Multi-frame} & 12/12 & 12/12 & 12/12 & 12/12 & {\color[HTML]{000000} \textbf{100\%}} & 11/12 & 12/12 & 12/12 & 12/12 & 11/12 & 11/12 & 12/12 & 6/12 & {\color[HTML]{000000} \textbf{91.7\%}} \\
\textbf{Multi-frame$^{a}$} & 12/12 & 12/12 & 12/12 & 12/12 & {\color[HTML]{000000} \textbf{100\%}} & 12/12 & 12/12 & 12/12 & 12/12 & 12/12 & 11/12 & 10/12 & 9/12 & {\color[HTML]{000000} \textbf{94.8\%}} \\ \hline
 & \multicolumn{14}{c}{Insertion efficiency (s)} \\ \hline
E2E vision-RL & 5$\pm$1.6 & 5$\pm$0.9 & 5$\pm$1.7 & 4$\pm$1.2 & 5$\pm$1.4 & 12$\pm$2.1 & 6$\pm$2.0 & $\times$ & $\times$ & $\times$ & $\times$ & $\times$ & 9$\pm$2.1 & 8$\pm$3.0 \\
E2E vision-RL$^{a}$ & 5$\pm$1.6 & 5$\pm$0.9 & 5$\pm$1.7 & 4$\pm$1.3 & 5$\pm$1.4 & 9$\pm$4.0 & 7$\pm$2.2 & $\times$ & $\times$ & $\times$ & $\times$ & $\times$ & $\times$ & 8$\pm$3.0 \\
Single-frame & 7$\pm$2.5 & 4$\pm$1.3 & 5$\pm$1.6 & 5$\pm$1.6 & {\color[HTML]{000000} 6$\pm$3.0} & 9$\pm$3.8 & 6$\pm$2.3 & 5$\pm$2.6 & 5$\pm$2.2 & 11$\pm$4.4 & 8$\pm$3.1 & 7$\pm$3.1 & 15$\pm$3.4 & {\color[HTML]{000000} 8$\pm$4.4} \\
\textbf{Single-frame$^{a}$} & 3$\pm$1.4 & 3$\pm$1.4 & 3$\pm$1.7 & 3$\pm$1.9 & {\color[HTML]{000000} \textbf{3$\pm$1.6}} & 7$\pm$5.1 & 3$\pm$1.8 & 4$\pm$2.1 & 3$\pm$1.8 & 5$\pm$3.1 & 4$\pm$4.1 & 4$\pm$4.1 & 4$\pm$1.4 & {\color[HTML]{000000} \textbf{4$\pm$3.3}} \\
\textbf{Multi-frame} & 4$\pm$1.0 & 3$\pm$0.9 & 3$\pm$0.8 & 3$\pm$0.8 & {\color[HTML]{000000} \textbf{3$\pm$1.0}} & 4$\pm$1.6 & 4$\pm$1.2 & 7$\pm$3.6 & 4$\pm$0.8 & 6$\pm$3.0 & 6$\pm$3.4 & 7$\pm$3.4 & 10$\pm$4.7 & {\color[HTML]{000000} \textbf{6$\pm$3.2}} \\
\textbf{Multi-frame$^{a}$} & 2$\pm$1.2 & 3$\pm$1.4 & 3$\pm$1.0 & 3$\pm$1.3 & {\color[HTML]{000000} \textbf{3$\pm$1.2}} & 3$\pm$2.5 & 4$\pm$1.7 & 4$\pm$4.8 & 3$\pm$1.6 & 6$\pm$2.8 & 4$\pm$2.4 & 4$\pm$3.4 & 3$\pm$0.5 & {\color[HTML]{000000} \textbf{4$\pm$3.0}} \\ \hline
\end{tabular}
\begin{tablenotes}
\item[\scalebox{1.0}{$\times$}] This superscript denotes the method is not general to the insertion tasks.
\item Cases with $^a$ are tested in eye-in-hand configuration, whilst without $^a$ are eye-to-hand configuration.
\item Boldface denotes the proposed methods achieved better results than the compared methods.
\end{tablenotes}
\end{threeparttable}
\label{efficiency_tab}
\end{table*}

Since the performance gap is the largest among the compared methods when the tolerance is 1$mm$ as shown in Fig.~\ref{success_rate}, we further compare the success rate and insertion efficiency respectively with 1$mm$ hole tolerance. The insertion efficiency is calculated by averaging over the total moving steps of the successful insertion trajectories. Cases that exceed the predefined safe region are regarded as a fail trajectory. The initial position error is 10$mm$ and the rotation error is $10^\circ$. 
The results of each shape are illustrated in Tab. \ref{efficiency_tab}. We find that our proposed multi-frame and single-frame policies achieve better averaged results on both success rate and insertion efficiency compared with the VSN policy and the E2E Vision-RL. Notably, they outperform the E2E Vision-RL by large margins in terms of success rate on the unseen shapes~(\textgreater90\% vs. \textless10\%). We also find that the multi-frame policy achieves faster insertion with fewer steps compared with the single-frame policy on both the seen and the unseen shapes and under both the eye-in-hand and eye-to-hand configurations. We believe that by considering the historical visual information and actions, the multi-frame policy is aware of the sensor prediction error and forges ahead bravely, while the single-frame policy can only proceed cautiously with small steps by only taking as input the current visual feature with uncertainty. 

\subsection{Real World Experiments}
\label{real_world}
We use the peg-hole pairs with around 0.6$mm$ tolerance for all the real-world experiments. We design the 3D mesh of the peg-hole pairs in Blender, an open software for 3D modeling, with the clearance between 0.5$mm$ to 0.7$mm$. The 3D print error is 0.2$mm$.

\begin{table*}[]
\centering
\setlength{\tabcolsep}{2.5pt}
\caption{Insertion Performance with 0.6$mm$ Tolerance in Real World}
\begin{threeparttable}[t]
\begin{tabular}{ccccccccccccccc}
\hline
 & \multicolumn{5}{c}{Seen} & \multicolumn{8}{c}{Unseen shape} &  \\
\multirow{-2}{*}{Methods} & 1 & 2 & 3 & 4 & Average & 1 & 2 & 3 & 4 & 5 & 6 & 7 & 8 & Average \\ \hline
\textbf{} & \multicolumn{14}{c}{5mm / Success rate} \\ \hline
Spiral Search $^{b}$ & 10/12 & 11/12 & 11/12 & 10/12 & 87.5\% & 9/12 & 9/12 & 10/12 & 9/12 & 8/12 & 8/12 & 7/12 & 7/12 & 69.7\% \\
E2E Force-RL $^{b}$ & 11/12 & 10/12 & 11/12 & 11/12 & 89.6\% & 9/12 & 9/12 & 8/12 & 9/12 & 8/12 & 8/12 & 5/12 & 5/12 & 65.3\% \\ \hline
E2E Vision-RL & 12/12 & 12/12 & 11/12 & 11/12 & 97.9\% & $\times$ & $\times$ & $\times$ & $\times$ & $\times$ & $\times$ & $\times$ & $\times$ & $\times$ \\
E2E Vision-RL$^{a}$ & 12/12 & 12/12 & 11/12 & 12/12 & 97.9\% & $\times$ & $\times$ & $\times$ & $\times$ & $\times$ & $\times$ & $\times$ & $\times$ & $\times$ \\
\textbf{Ours} & 11/12 & 11/12 & 10/12 & 11/12 & {\color[HTML]{000000} 89.6\%} & 9/12 & 9/12 & 8/12 & 7/12 & 7/12 & 7/12 & 6/12 & 6/12 & {\color[HTML]{000000} \textbf{61.5\%}} \\
\textbf{Ours$^{a}$} & 11/12 & 10/12 & 12/12 & 11/12 & {\color[HTML]{000000} 91.75} & 9/12 & 10/12 & 9/12 & 7/12 & 7/12 & 7/12 & 7/12 & 7/12 & {\color[HTML]{000000} \textbf{65.6\%}} \\ \hline
 & \multicolumn{14}{c}{5mm / Insertion efficiency (s)} \\ \hline
Spiral Search $^{b}$ & 10$\pm$3.1 & 9$\pm$3.0 & 8$\pm$2.5 & 10$\pm$2.8 & 9$\pm$3.0 & 10$\pm$3.2 & 9$\pm$3.4 & 11$\pm$3.2 & 10$\pm$3.7 & 11$\pm$3.5 & 9$\pm$3.1 & 11$\pm$3.8 & 10$\pm$3.3 & 10$\pm$3.3 \\
E2E Force-RL $^{b}$ & 4$\pm$1.6 & 3$\pm$1.1 & 5$\pm$1.3 & 3$\pm$2.1 & 4$\pm$1.7 & 4$\pm$0.9 & 5$\pm$1.2 & 3$\pm$0.9 & 4$\pm$1.0 & 3$\pm$1.0 & 5$\pm$1.1 & 11$\pm$2.3 & 6$\pm$1.5 & 5$\pm$1.8 \\ \hline
E2E Vision-RL & 5$\pm$0.9 & 3$\pm$1.1 & 3$\pm$1.3 & 4$\pm$1.6 & 4$\pm$1.2 & $\times$ & $\times$ & $\times$ & $\times$ & $\times$ & $\times$ & $\times$ & $\times$ & $\times$ \\
E2E Vision-RL$^{a}$ & 4$\pm$0.8 & 4$\pm$1.2 & 3$\pm$0.9 & 3$\pm$1.1 & 5$\pm$1.1 & $\times$ & $\times$ & $\times$ & $\times$ & $\times$ & $\times$ & $\times$ & $\times$ & $\times$ \\
\textbf{Ours} & 4$\pm$1.5 & 4$\pm$0.8 & 5$\pm$1.7 & 5$\pm$1.2 & {\color[HTML]{000000} \textbf{4$\pm$1.1}} & 3$\pm$1.7 & 4$\pm$1.5 & 4$\pm$1.5 & 5$\pm$2.0 & 4$\pm$1.5 & 5$\pm$2.2 & 6$\pm$1.9 & 5$\pm$1.4 & {\color[HTML]{000000} \textbf{4$\pm$1.5}} \\
\textbf{Ours$^{a}$} & 3$\pm$0.4 & 4$\pm$0.8 & 3$\pm$1.3 & 4$\pm$1.4 & {\color[HTML]{000000} \textbf{3$\pm$1.2}} & 4$\pm$1.5 & 3$\pm$1.7 & 3$\pm$0.9 & 4$\pm$1.5 & 4$\pm$1.3 & 4$\pm$1.6 & 6$\pm$1.5 & 5$\pm$1.5 & {\color[HTML]{000000} \textbf{4$\pm$1.5}} \\ \hline
 & \multicolumn{14}{c}{10mm / Success rate} \\ \hline
Spiral Search $^{b}$ & 5/12 & 6/12 & 5/12 & 5/12 & 43.8\% & 6/12 & 7/12 & 6/12 & 5/12 & 5/12 & 4/12 & 1/12 & 3/12 & 36.5\% \\
E2E Force-RL $^{b}$ & $\times$ & $\times$ & $\times$ & $\times$ & $\times$ & $\times$ & $\times$ & $\times$ & $\times$ & $\times$ & $\times$ & $\times$ & $\times$ & $\times$ \\ \hline
E2E Vision-RL & 11/12 & 12/12 & 11/12 & 11/12 & 93.8\% & $\times$ & $\times$ & $\times$ & $\times$ & $\times$ & $\times$ & $\times$ & $\times$ & $\times$ \\
E2E Vision-RL$^{a}$ & 10/12 & 11/12 & 12/12 & 11/12 & 91.7\% & $\times$ & $\times$ & $\times$ & $\times$ & $\times$ & $\times$ & $\times$ & $\times$ & $\times$ \\
\textbf{Ours} & 11/12 & 11/12 & 10/12 & 10/12 & {\color[HTML]{343434} 87.5\%} & 7/12 & 8/12 & 7/12 & 6/12 & 7/12 & 6/12 & 6/12 & 4/12 & {\color[HTML]{000000} \textbf{53.1\%}} \\
\textbf{Ours$^{a}$} & 10/12 & 11/12 & 11/12 & 11/12 & {\color[HTML]{000000} 89.6\%} & 8/12 & 8/12 & 7/12 & 6/12 & 6/12 & 6/12 & 6/12 & 4/12 & {\color[HTML]{000000} \textbf{54.3\%}} \\ \hline
 & \multicolumn{14}{c}{10mm / Insertion efficiency (s)} \\ \hline
Spiral Search $^{b}$ & 23$\pm$4.5 & 25$\pm$4.6 & 21$\pm$3.9 & 23$\pm$3.6 & 23$\pm$4.1 & 22$\pm$4.8 & 22$\pm$4.4 & 23$\pm$4.3 & 24$\pm$4.4 & 23$\pm$5.1 & 24$\pm$4.5 & 25$\pm$3.1 & 25$\pm$4.5 & 24$\pm$4.5 \\
E2E Force-RL $^{b}$ & $\times$ & $\times$ & $\times$ & $\times$ & $\times$ & $\times$ & $\times$ & $\times$ & $\times$ & $\times$ & $\times$ & $\times$ & $\times$ & $\times$ \\ \hline
E2E Vision-RL & 5$\pm$1.3 & 5$\pm1.8$ & 6$\pm$2.4 & 4$\pm$1.5 & 5$\pm$1.6 & $\times$ & $\times$ & $\times$ & $\times$ & $\times$ & $\times$ & $\times$ & $\times$ & $\times$ \\
E2E Vision-RL$^{a}$ & 5$\pm$1.2 & 4$\pm$1.5 & 5$\pm$1.7 & 4$\pm$1.2 & 5$\pm$1.5 & $\times$ & $\times$ & $\times$ & $\times$ & $\times$ & $\times$ & $\times$ & $\times$ & $\times$ \\
\textbf{Ours} & 4$\pm$0.9 & 5$\pm$1.2 & 6$\pm$1.0 & 5$\pm$1.6 & {\color[HTML]{000000} \textbf{5$\pm$1.3}} & 5$\pm$1.8 & 5$\pm$1.2 & 6$\pm$1.3 & 5$\pm$1.5 & 5$\pm$1.5 & 4$\pm$1.3 & 6$\pm$2.4 & 4$\pm$1.7 & {\color[HTML]{000000} \textbf{5$\pm$1.5}} \\
\textbf{Ours$^{a}$} & 4$\pm$0.4 & 5$\pm$1.1 & 5$\pm$0.8 & 4$\pm$1.5 & {\color[HTML]{000000} \textbf{4$\pm$1.2}} & 5$\pm$2.0 & 4$\pm$1.3 & 4$\pm$1.4 & 4$\pm$1.7 & 5$\pm$1.3 & 4$\pm$0.8 & 6$\pm$1.9 & 5$\pm$1.3 & {\color[HTML]{000000} \textbf{5$\pm$1.4}} \\ \hline
\end{tabular}
\begin{tablenotes}
\item[\scalebox{1.0}{$\times$}] This superscript denotes the method is not general to the insertion tasks.
\item Cases with $^a$ are tested in eye-in-hand configuration, whilst without $^a$ are eye-to-hand configuration.
\item Cases with $^b$ denote the 3-DoF peg-in-hole.
\item Boldface denotes the proposed methods achieved better results than the compared methods.
\end{tablenotes}
\end{threeparttable}
\label{comparison_tab}
\end{table*}

\subsubsection{Peg-in-hole with 0.6\textit{mm} Tolerance}
We compare the proposed multi-frame policy with three previous works on success rate, insertion efficiency (defined in Sec.~\ref{submm}), and unseen shape generalization. The compared alternatives include the previously mentioned E2E Vision-RL and the following two methods. 

\textbf{Spiral Search}~\cite{2018Multi}: 
Will move downward in the $z$-axis until a force limit is reached, indicating the peg has been in contact with the hole. Then the peg moves outwards in a spiral while pressing against the hole surface. When the peg moves over the hole within a success region, the force in $z$-axis will help press the peg to insert compliantly. Otherwise, the peg will move until exceeding the uncertainty boundary and fail.

\textbf{E2E Force-RL}~\cite{inoue2017deep}:
Learns the policy end-to-end, which outputs the 2D translate displacement $[dx, dy]$ in the horizon space by taking the 6D force/torque $[F_x, F_y, F_z, M_x, M_y, M_z]$ as input. Force control is performed in the $z$-axis to maintain a constant contact as with the spiral search. The sparse reward is given when the insert successes within possibly the minimum steps as defined in \cite{inoue2017deep}. The model is trained in simulation and transferred to the real-world directly.

As shown in Tab. \ref{comparison_tab}, the force-based methods like spiral search and E2E Force-RL achieve comparable insertion success rate with the vision-based methods when the initial error is within 5$mm$. However, spiral search is inefficient and the success rate drops rapidly with a larger initial error (10$mm$), as the search-based open-loop control method relies heavily on the initial position. For the E2E Force-RL, it fails to be general when the initial error is 10$mm$. We believe that the force/torque distribution gap between the simulation and the real-world grows wider with a larger initial error. And directly policy transfer finds it hard to tolerate the larger gap, which results in poor performance. Moreover, in experiments we found the force-based methods fail to learn the rotation alignment under large yaw error in $z$-axis (10$^{\circ}$). Due to this reason, we take one step back by performing 3-DoF peg-in-hole that only outputs $dx, dy, dz$ in the action space, which is unfair to the vision-based 4-DoF peg-in-hole that considers the extra $z$-axis rotation error. For the vision-based methods, we find that the E2E Vision-RL achieves both a high success rate and efficiency by over-fitting to the seen shapes during training, but pays for the price for not being general to the unseen shapes. Our proposed multi-frame policy can realize both a higher success rate and efficiency while generalizing to the unseen shapes under both the eye-in-hand and the eye-to-hand configurations. 

\subsubsection{Perception Adaptation}
\begin{figure}[t]
\centering
\includegraphics[width=0.48\textwidth]{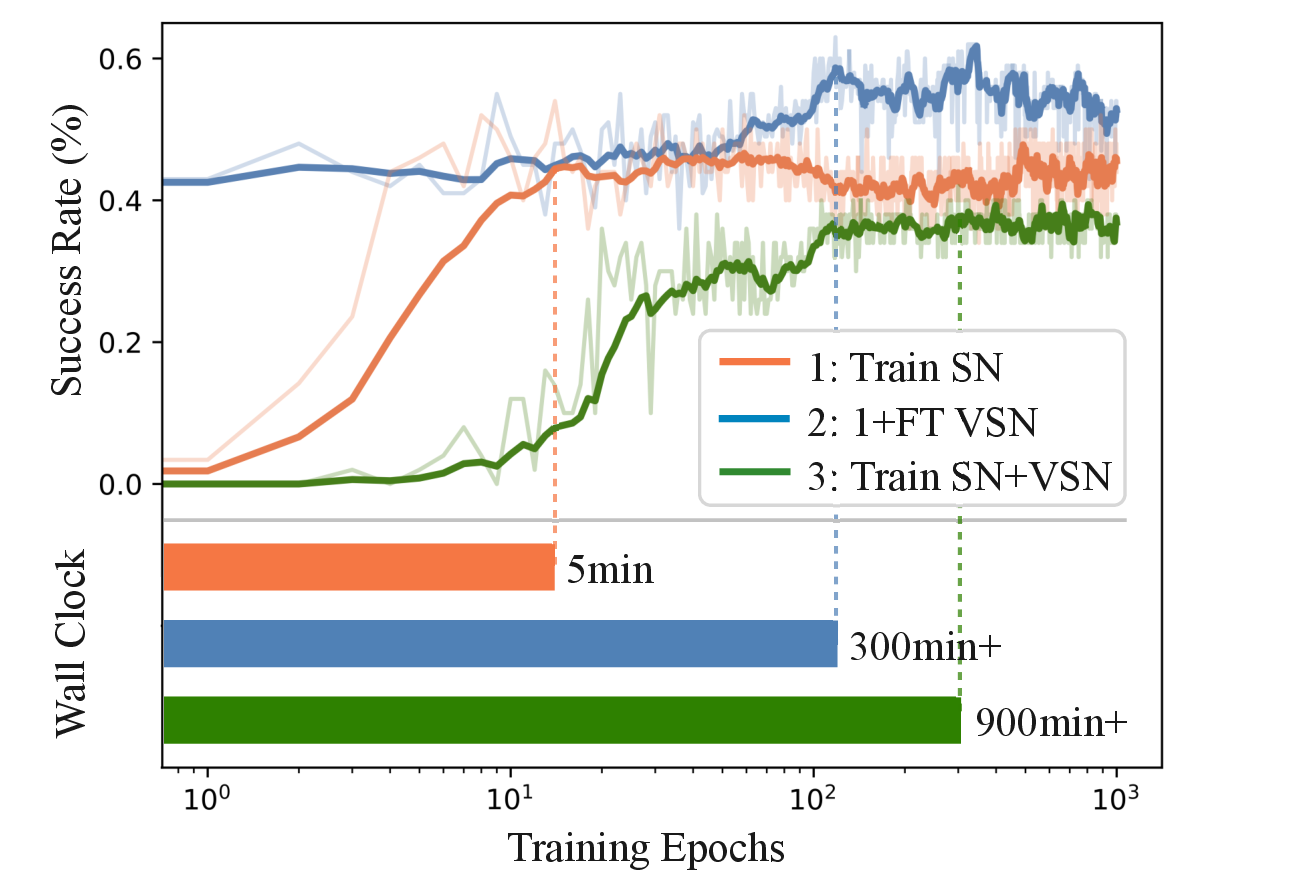}
\caption{Success rate of the VSN policy with three different adaptation strategies.}
\label{sim2real_curve}
\end{figure}

We conduct the perception adaptation to achieve sim-to-real using efficient supervised learning. The training data is collected and annotated in the real-world platform automatically. Three adaptation strategies are compared. The first strategy only trains the SN and keeps the VSN fixed which is pre-trained in simulation. The second one begins with a trained SN and fine-tunes the VSN pre-trained in simulation. The third one trains the SN and the VSN from scratch with real-world data. 500 RGB images with annotations for both the SN and the VSN are collected in around 15 minutes by randomly changing the robot end-effector's pose relative to the hole without contact. The relative distance ranges from -10 to 10$mm$ in the $x$ and $y$ axis, and the relative $z$-axis yaw rotation ranges from $-10^{\circ}$ to $10^\circ$. The 500 images are divided into the training set and the testing set by a ratio of 9:1. Fig. \ref{sim2real_curve} illustrates the success rate of the VSN policy on the testing set. We can find that the performance curve of the first strategy converges fast ( $\sim$10 training epochs) in around 5 minutes, which achieves 0.97 $MeanIoU$\cite{long2015fully} with nearly perfect segmentation precision and around 0.45 success rate. 
The success rate of the second strategy outperforms the first one by up to 10\%. Significant improvement of the VSN is achieved by fine-tuning the VSN with real-world data, though at the cost of around 5 hours' training.
The curve of the third strategy converges until 300 epochs, which takes around 15 hours for training. The final performance of the third strategy does not exceed that of the first strategy, which we believe is because of the lack of real-world data for training the VSN from scratch. In real-world experiments, we adopt the first strategy for fast perception adaptation.

\subsubsection{Robustness under Vision Occlusion} 
We evaluate the robustness of the proposed system when the vision occlusion occurs with the Fillet-1 shape (see Fig. \ref{setting}). Fig. \ref{occlusion} shows the perspective of the camera, where the hole is occluded entirely by the peg in the initial state. Under the vision occlusion, the VSN policy loses the seam as reference for alignment, thus drifting off the safe zone where no historical or future information can provide the reference for the policy to pull back. However, both the multi-frame and single-frame policy can recover from the vision blind with the help of the CN, but the multi-frame policy recovers faster than the single-frame policy. 
We infer that the prediction uncertainty from the VSN is reduced by considering the historical sensor features and actions, which contributes to the convergence of the multi-frame policy.

\subsection{EV Charging}
Finally, we conduct the real-world experiments in the automatic EV charging system, which is a complete unseen insertion scenario, to evaluate the practicability of our proposed algorithms. We achieve fast perception adaptation on the platform within 20 minutes, including automatic data collection, annotation, and segmentation training. The policy is trained with the same four seen shapes in simulation as mentioned above and developed in the vehicle charging case directly without fine-tuning. The initial position and rotation error is within 10$mm$ and $10^\circ$ respectively.
\subsubsection{Static Insertion}
The experiment platform is built with the Tesla Model 3, the UR5e robot arm, the RealSense D435 camera, the charging plug, and the LED flash as shown in Fig. \ref{vehicle}(a). The camera is fixed at the robot's wrist with the eye-in-hand configuration. The policy starts by commanding the robot to approach the hole, and touch the surface with constant contact based on force signals from the build-in F/T sensor. Then the pose alignment is achieved by the proposed algorithm. Finally, when the aligning process finishes, the peg will be pushed into the hole compliantly under the constant force to realize the insertion. We count a 10/10 success rate under random initial poses, with each insertion completed within 3s, which verifies the practicability of the proposed system.
\subsubsection{Dynamic Insertion}
We further evaluate the proposed framework in more challenging conditions. The agent needs to perform a dynamic insertion while a person is manually moving the EV socket (see Fig.~\ref{vehicle}(b)). This is much harder than the static insertions.
Results show that our methods can solve these very challenging tasks effectively.
More insertion cases can be found in the supplemented videos~\cite{xiel}.

\subsection{Discussion}
Through the simulations and real-world experiments, we answer the previously mentioned five questions on generalization, efficiency, precision, robustness, and adaptation for peg-in-hole, and discuss the strengths and potentials of our proposed framework.
\begin{itemize}
\item [1)]
The experiments examine the effectiveness and practicability of our proposed framework across various peg insertion tasks. Evaluation of a series of hole shapes demonstrates the simulation-based generic policy can apply to a wide variety of unseen insertion tasks, which proves that the policy is \emph{trained with several shapes, and tested to another set of shapes unseen in the training set.}

\item [2)]
Our approach achieves efficient insertion under large initial error (10$mm$ and 10$^{\circ}$) in around 3s, which is much faster than the force-based methods. The error range almost covers the initial error caused by the upstream tasks such as object pose estimation, mobile robot navigation, etc. However, theoretically, by enlarging the measurement range or resolution of the VSN, our method can scale up to a broader error range as long as the hole is in the visual field. 

\begin{figure}[t]
\centering
\includegraphics[height=58mm,width=0.5\textwidth]{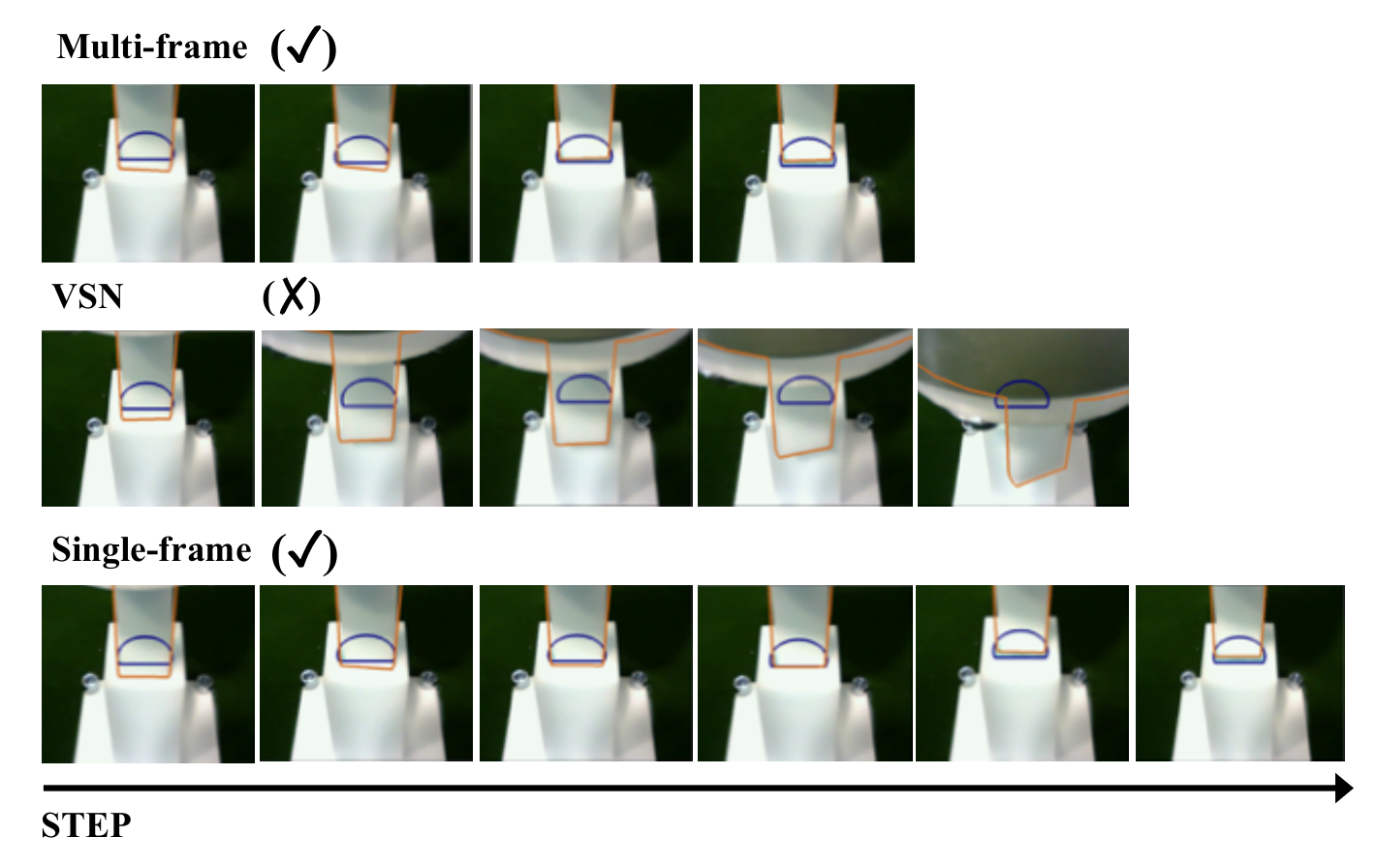}
\caption{Peg-hole pose aligning trajectory under camera view: the multi-frame policy performs fast and accurate pose alignment; the VSN policy drifts away from the safe zone; the single-frame policy can achieve pose alignment but is inefficient.}
\label{occlusion}
\end{figure}

\item [3)]
Experiments in both simulation and real-world prove that under 1$mm$ hole tolerance, our approach achieves a high success rate (\textgreater~90\%) and insertion efficiency on both the seen and unseen holes, and it performs well in the dynamic insertion scenario, which is a more challenge task. We further explore the performance of our approach with 0.6$mm$ tolerance in the real world. The insertion success rate remains $\sim$~90\% with tighter tolerances on the seen holes, while achieving \textgreater~50\% success rate on the unseen holes, which is consistent with the simulated results.
To improve the performance on the unseen holes under sub-$mm$ tolerance, we can use extra task-specific data to fine-tune the VSN structure. We prove more than 10\% improvement in the success rate can be achieved for the VSN with extra training. We do not quantify the improvement for the peg-in-hole by fine-tuning the VSN in the real-world experiments, as a follow-up force-based method can be combined to cover the performance gap if sub-$mm$ precision is required.

\begin{figure*}[t]
\centering
\includegraphics[height=130mm,width=1\textwidth]{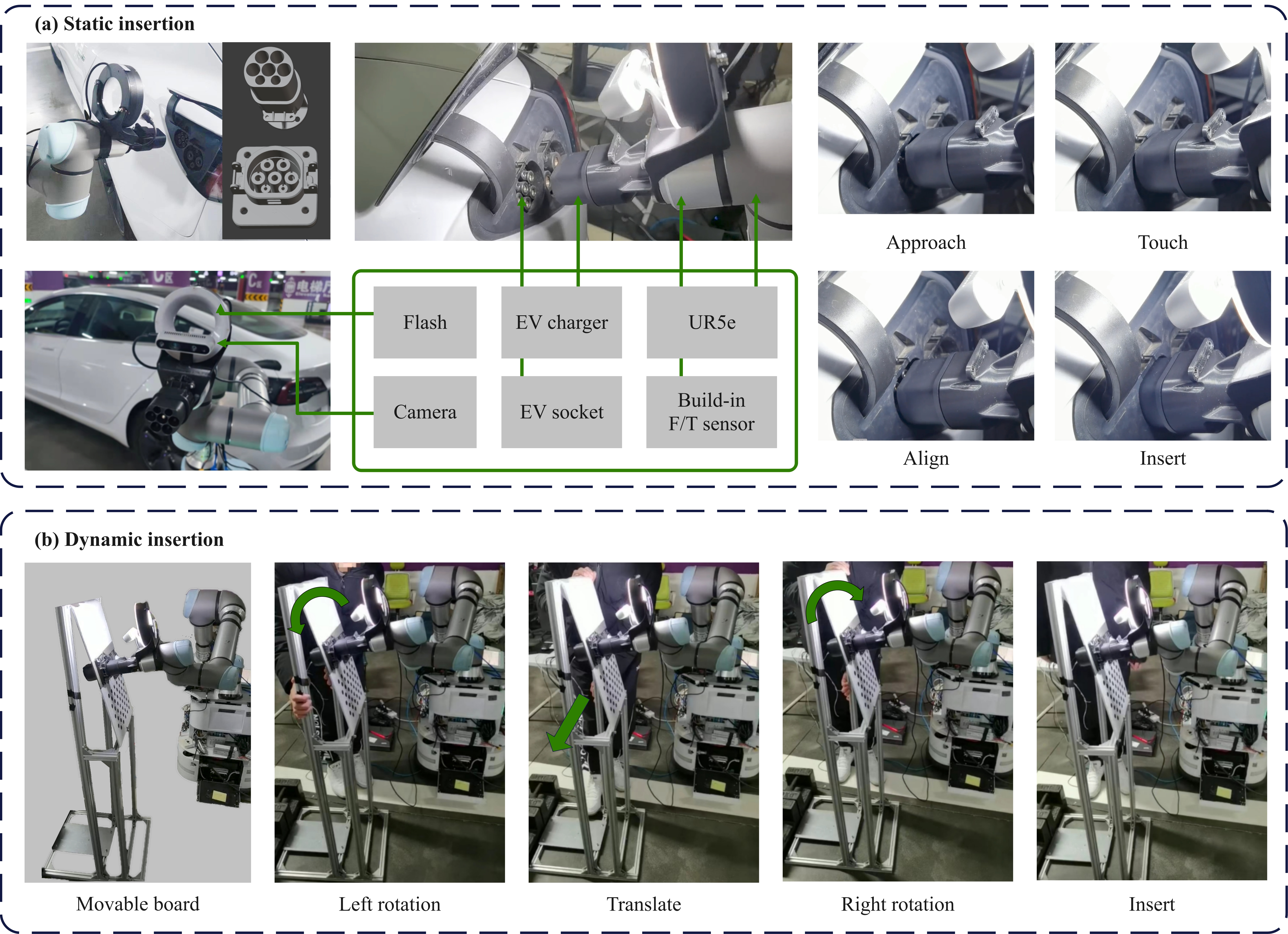}
\caption{Real-world plugging of the EV charging system. (a) Static insertion in a real car of Tesla Model 3. (b) Dynamic insertion with the EV socket mounted to a movable board.}
\label{vehicle}
\end{figure*}

\item [4)]
Under vision occlusion circumstances where the hole is totally occluded, the VSN policy faces the ambiguity problem as a given visual state can correspond to multiple pose initialization. Thus the VSN policy fails to align the peg-hole pose when the vision occlusion occurs. 
The proposed single-frame and multi-frame policy solve the ambiguity by considering future or historical information with the CN module. And the multi-frame policy proves to be more efficient for peg insertion tasks in the real-world experiments.
Notably, the multi-frame policy tasks as input the image sequence, thus occupying more graphics memory than the single-frame policy. In lightweight applications where computing resources are limited, the single-frame policy can be a competitive alternative method.


\item [5)]
By fine-tuning the SN, our framework can adapt to various real-world insertion tasks with only hundreds of images in 20 minutes, including data collection, annotation, and model training. The adaptation only needs one-minute human teaching with an accurate insertion to achieve automatic data collection and annotation, without the requirement of expert programming.
Similar to the traditional position control for peg insertion, which calls for an accurate insertion by a human to achieve the repetitive positioning, our method requires the same \emph{minimal additional cost}.

\item [6)]
While our framework presents a step toward generalizable peg insertion, the generalization of visual perception remains a great challenge, especially in the dynamic and unstructured outdoor environments. Combining the current domain randomization techniques with our proposed automatic data collection and annotation pipeline to acquire a huge diversity of data would be a promising direction for future research.

\end{itemize}

\section{Conclusion}
\label{dis}
In the peg insertion tasks, vision feedback control is preferred compared with other input modalities (e.g. force, tactile, senseless, etc.), especially under large initial error. However, current vision-based solutions for peg-in-hole show poor performance to new objects with unseen shape.
In this article, we present a vision feedback framework for precise peg-in-hole towards unseen shape generalization. Our framework enables training a generic insertion policy with several shapes in simulation, and adapting to arbitrary unseen shapes in real-world with minimal sim-to-real cost. The cost contains a fast perception adaptation process, which calls for one-minute manual teaching to bootstrap the automatic data collection and annotation pipeline. With the fast adaptable-adaptation module, the simulation-based generic policy can quickly generalize to various real-world insertion tasks, either under the eye-to-hand or eye-in-hand configuration. 
In the future work, 
we plan to pursue the tight insertion tasks for higher precision and robustness by combining the vision and force/torque data and extend to the 6-DoF insertion tasks.
We also want to explore the use of new progress in computer vision fields, such as domain randomization, to achieve perception generalization.

\begin{acknowledgments}
This work was supported by National Natural Science Foundation of China (62173293), Zhejiang Provincial Natural Science Foundation of China (LD22E050007), and the Fundamental Research Funds for Science and Technology on Space Intelligent Control Laboratory (2021-JCJQ-LB-010-13).
\end{acknowledgments}
\section*{AUTHOR DECLARATIONS}

\subsection*{Conflict of Interest}
The authors have no conflicts to disclose.

\subsection*{Author Contributions}
\noindent$\mathbf{Liang~Xie}$: Conceptualization (equal); Methodology (equal); Software (equal); Validation (equal); Writing – original draft (equal). $\mathbf{Hongxiang~Yu}$: Investigation (equal). 
$\mathbf{Kechun~Xu}$: Investigation (equal). 
$\mathbf{Tong~Yang}$: Investigation (equal). 
$\mathbf{MinHang~Wang}$: Resources (equal).
$\mathbf{Haojian~Lu}$: Funding acquisition (equal).
$\mathbf{Rong~Xiong}$: Supervision (equal).
$\mathbf{Yue~Wang}$: Writing - review $\&$ editing  (equal).
\section*{Data Availability Statement}

The data that support the findings of this study are available from the corresponding author upon reasonable request.

\nocite{*}
\bibliography{aipsamp}

@article{haugaard2020fast,
  title={Fast robust peg-in-hole insertion with continuous visual servoing},
  author={Haugaard, Rasmus Laurvig and Langaa, Jeppe and Sloth, Christoffer and Buch, Anders Glent},
  journal={arXiv preprint arXiv:2011.06399},
  year={2020}
}

@article{morgan2021vision,
  title={Vision-driven compliant manipulation for reliable, high-precision assembly tasks},
  author={Morgan, Andrew S and Wen, Bowen and Liang, Junchi and Boularias, Abdeslam and Dollar, Aaron M and Bekris, Kostas},
  journal={arXiv preprint arXiv:2106.14070},
  year={2021}
}

@inproceedings{lee2019making,
  title={Making sense of vision and touch: Self-supervised learning of multimodal representations for contact-rich tasks},
  author={Lee, Michelle A and Zhu, Yuke and Srinivasan, Krishnan and Shah, Parth and Savarese, Silvio and Fei-Fei, Li and Garg, Animesh and Bohg, Jeannette},
  booktitle={2019 International Conference on Robotics and Automation (ICRA)},
  pages={8943--8950},
  year={2019},
  organization={IEEE}
}

@article{dong2021tactile,
  title={Tactile-RL for Insertion: Generalization to Objects of Unknown Geometry},
  author={Dong, Siyuan and Jha, Devesh K and Romeres, Diego and Kim, Sangwoon and Nikovski, Daniel and Rodriguez, Alberto},
  journal={arXiv preprint arXiv:2104.01167},
  year={2021}
}

@article{ding2019transferable,
  title={Transferable force-torque dynamics model for peg-in-hole task},
  author={Ding, Junfeng and Wang, Chen and Lu, Cewu},
  journal={arXiv preprint arXiv:1912.00260},
  year={2019}
}

@inproceedings{inoue2017deep,
  title={Deep reinforcement learning for high precision assembly tasks},
  author={Inoue, Tadanobu and De Magistris, Giovanni and Munawar, Asim and Yokoya, Tsuyoshi and Tachibana, Ryuki},
  booktitle={2017 IEEE/RSJ International Conference on Intelligent Robots and Systems (IROS)},
  pages={819--825},
  year={2017},
  organization={IEEE}
}

@article{stevvsic2020learning,
  title={Learning to assemble: Estimating 6D poses for robotic object-object manipulation},
  author={Stev{\v{s}}i{\'c}, Stefan and Christen, Sammy and Hilliges, Otmar},
  journal={IEEE Robotics and Automation Letters},
  volume={5},
  number={2},
  pages={1159--1166},
  year={2020},
  publisher={IEEE}
}

@inproceedings{zakka2020form2fit,
  title={Form2fit: Learning shape priors for generalizable assembly from disassembly},
  author={Zakka, Kevin and Zeng, Andy and Lee, Johnny and Song, Shuran},
  booktitle={2020 IEEE International Conference on Robotics and Automation (ICRA)},
  pages={9404--9410},
  year={2020},
  organization={IEEE}
}

@inproceedings{triyonoputro2019quickly,
  title={Quickly inserting pegs into uncertain holes using multi-view images and deep network trained on synthetic data},
  author={Triyonoputro, Joshua C and Wan, Weiwei and Harada, Kensuke},
  booktitle={2019 IEEE/RSJ International Conference on Intelligent Robots and Systems (IROS)},
  pages={5792--5799},
  year={2019},
  organization={IEEE}
}

@inproceedings{mnih2016asynchronous,
  title={Asynchronous methods for deep reinforcement learning},
  author={Mnih, Volodymyr and Badia, Adria Puigdomenech and Mirza, Mehdi and Graves, Alex and Lillicrap, Timothy and Harley, Tim and Silver, David and Kavukcuoglu, Koray},
  booktitle={International conference on machine learning},
  pages={1928--1937},
  year={2016},
  organization={PMLR}
}

@inproceedings{schoettler2020deep,
  title={Deep reinforcement learning for industrial insertion tasks with visual inputs and natural rewards},
  author={Schoettler, Gerrit and Nair, Ashvin and Luo, Jianlan and Bahl, Shikhar and Ojea, Juan Aparicio and Solowjow, Eugen and Levine, Sergey},
  booktitle={2020 IEEE/RSJ International Conference on Intelligent Robots and Systems (IROS)},
  pages={5548--5555},
  year={2020},
  organization={IEEE}
}

@misc{tobin2017domain,
      title={Domain Randomization for Transferring Deep Neural Networks from Simulation to the Real World}, 
      author={Josh Tobin and Rachel Fong and Alex Ray and Jonas Schneider and Wojciech Zaremba and Pieter Abbeel},
      year={2017},
      eprint={1703.06907},
      archivePrefix={arXiv},
      primaryClass={cs.RO}
}

@misc{long2015fully,
      title={Fully Convolutional Networks for Semantic Segmentation}, 
      author={Jonathan Long and Evan Shelhamer and Trevor Darrell},
      year={2015},
      eprint={1411.4038},
      archivePrefix={arXiv},
      primaryClass={cs.CV}
}

@article{2019Compare,
  title={Compare Contact Model-based Control and Contact Model-free Learning: A Survey of Robotic Peg-in-hole Assembly Strategies},
  author={ Xu, J.  and  Hou, Z.  and  Liu, Z.  and H Qiao},
  year={2019},
}

@article{2021Offline,
  title={Offline Meta-Reinforcement Learning for Industrial Insertion},
  author={ Zhao, T. Z.  and  Luo, J.  and  Sushkov, O.  and  Pevceviciute, R.  and  Heess, N.  and  Scholz, J.  and  Schaal, S.  and  Levine, S. },
  year={2021},
}

@article{2021Robust,
  title={Robust Multi-Modal Policies for Industrial Assembly via Reinforcement Learning and Demonstrations: A Large-Scale Study},
  author={ Luo, J.  and  Sushkov, O.  and  Pevceviciute, R.  and  Lian, W.  and  Scholz, J. },
  year={2021},
}

@article{2018Numerical,
  title={Numerical Coordinate Regression with Convolutional Neural Networks},
  author={ Nibali, A.  and  He, Z.  and  Morgan, S.  and  Prendergast, L. },
  year={2018},
}

@article{xiel,
  title={Learning Visual Guided Policy for Precision
Peg-in-hole with Unseen Shape Generalization},
  author={Liang Xie},
  year={2022},
  url={https://github.com/xieliang555/SFN.git}
}

@article{2018Multi,
  title={Multi-Robot Assembly Strategies and Metrics},
  author={ Marvel, Jeremy A.  and  Bostelman, Roger  and  Falco, Joe },
  journal={ACM Computing Surveys},
  volume={51},
  number={1},
  pages={1-32},
  year={2018},
}

@article{2010Neural,
  title={Neural Synergy Between Kinetic Vision and Touch},
  author={ Blake, Randolph  and  Sobel, Kenith V.  and  James, Thomas W. },
  journal={Psychol},
  volume={15},
  number={6},
  pages={397-402},
  year={2010},
}

@article{2018Reinforcement,
  title={Reinforcement and Imitation Learning for Diverse Visuomotor Skills},
  author={ Zhu, Y.  and  Wang, Z.  and  Merel, J.  and  Rusu, A.  and  Heess, N. },
  year={2018},
}

@article{2018Learning,
  title={Learning Dexterous In-Hand Manipulation},
  author={ Andrychowicz, M.  and  Baker, B.  and  Chociej, M.  and  Jozefowicz, R.  and  Mcgrew, B.  and  Pachocki, J.  and  Petron, A.  and  Plappert, M.  and  Powell, G.  and  Ray, A. },
  year={2018},
}

@article{2017Sim,
  title={Sim-to-Real Transfer of Robotic Control with Dynamics Randomization},
  author={ Peng, X. B.  and  Andrychowicz, M.  and  Zaremba, W.  and  Abbeel, P. },
  year={2017},
}

@article{2018Using,
  title={Using Simulation and Domain Adaptation to Improve Efficiency of Deep Robotic Grasping},
  author={ Bousmalis, K.  and  Irpan, A.  and  Wohlhart, P.  and  Bai, Y.  and  Kelcey, M.  and  Kalakrishnan, M.  and  Downs, L.  and  Ibarz, J.  and  Pastor, P.  and  Konolige, K. },
  journal={IEEE},
  year={2018},
}

@INPROCEEDINGS{7576815,
  author={Tang, Te and Lin, Hsien-Chung and Zhao, Yu and Fan, Yongxiang and Chen, Wenjie and Tomizuka, Masayoshi},
  booktitle={2016 IEEE International Conference on Advanced Intelligent Mechatronics (AIM)}, 
  title={Teach industrial robots peg-hole-insertion by human demonstration}, 
  year={2016},
  volume={},
  number={},
  pages={488-494},
  doi={10.1109/AIM.2016.7576815}}

@article{2020Meta,
  title={Meta-Reinforcement Learning for Robotic Industrial Insertion Tasks},
  author={ Schoettler, G.  and  Nair, A.  and  Ojea, J. A.  and  Levine, S.  and  Solowjow, E. },
  year={2020},
}

@article{2015Learning2,
  title={Learning Contact-Rich Manipulation Skills with Guided Policy Search},
  author={ Levine, S.  and  Wagener, N.  and  Abbeel, P. },
  journal={Proceedings IEEE International Conference on Robotics and Automation},
  year={2015},
}

@article{2018Learning2,
  title={Learning Robotic Assembly from CAD},
  author={ Thomas, G.  and  Chien, M.  and  Tamar, A.  and  Ojea, J. A.  and  Abbeel, P. },
  pages={1-9},
  year={2018},
}

@article{2020Variable,
  title={Variable Compliance Control for Robotic Peg-in-Hole Assembly: A Deep Reinforcement Learning Approach},
  author={ Cristian, B. H.  and  Petit, D.  and  Ramirez-Alpizar, I. G.  and  Harada, K. },
  journal={Applied Sciences},
  volume={2020},
  number={10},
  pages={6923},
  year={2020},
}

@article{Zhang2019Jamming,
  title={Jamming Analysis and Force Control for Flexible Dual Peg-in-Hole Assembly},
  author={Zhang and Kuangen and Jing and Chen and Heping and Zhao and Jianguo and Ken},
  journal={IEEE Transactions on Industrial Electronics},
  volume={66},
  number={3},
  pages={1930-1939},
  year={2019},
}

@article{fei2003assembly,
  title={An assembly process modeling and analysis for robotic multiple peg-in-hole},
  author={Fei, Yanqiong and Zhao, Xifang},
  journal={Journal of Intelligent and Robotic Systems},
  volume={36},
  number={2},
  pages={175--189},
  year={2003},
  publisher={Springer}
}

@article{contact-rich,
  doi = {10.48550/ARXIV.1501.05611},
  
  url = {https://arxiv.org/abs/1501.05611},
  
  author = {Levine, Sergey and Wagener, Nolan and Abbeel, Pieter},
  
  title = {Learning Contact-Rich Manipulation Skills with Guided Policy Search},
  
  publisher = {arXiv},
  
  year = {2015},
  
  copyright = {arXiv.org perpetual, non-exclusive license}
}

@misc{efficient2021,
  doi = {10.48550/ARXIV.2105.04607},
  
  url = {https://arxiv.org/abs/2105.04607},
  
  author = {Endrawis, Shadi and Leibovich, Gal and Jacob, Guy and Novik, Gal and Tamar, Aviv},
  
  title = {Efficient Self-Supervised Data Collection for Offline Robot Learning},
  
  publisher = {arXiv},
  
  year = {2021},
  
  copyright = {arXiv.org perpetual, non-exclusive license}
}

@misc{affordance2019,
  doi = {10.48550/ARXIV.1903.04053},
  
  url = {https://arxiv.org/abs/1903.04053},
  
  author = {Hämäläinen, Aleksi and Arndt, Karol and Ghadirzadeh, Ali and Kyrki, Ville},
  
  title = {Affordance Learning for End-to-End Visuomotor Robot Control},
  
  publisher = {arXiv},
  
  year = {2019},
  
  copyright = {arXiv.org perpetual, non-exclusive license}
}

@misc{primitive2021,
  doi = {10.48550/ARXIV.2110.12618},
  
  url = {https://arxiv.org/abs/2110.12618},
  
  author = {Zhang, Xiang and Jin, Shiyu and Wang, Changhao and Zhu, Xinghao and Tomizuka, Masayoshi},
  
  title = {Learning Insertion Primitives with Discrete-Continuous Hybrid Action Space for Robotic Assembly Tasks},
  
  publisher = {arXiv},
  
  year = {2021},
  
  copyright = {arXiv.org perpetual, non-exclusive license}
}

@inproceedings{Puang_2020,
	doi = {10.1109/iros45743.2020.9341370},
  
	url = {https://doi.org/10.1109%2Firos45743.2020.9341370},
  
	year = 2020,
	month = {oct},
  
	publisher = {{IEEE}
},
  
	author = {En Yen Puang and Keng Peng Tee and Wei Jing},
  
	title = {{KOVIS}: Keypoint-based Visual Servoing with Zero-Shot Sim-to-Real Transfer for Robotics Manipulation},
  
	booktitle = {2020 {IEEE}/{RSJ} International Conference on Intelligent Robots and Systems ({IROS})}
}

@misc{zhu2018,
  doi = {10.48550/ARXIV.1802.09564},
  
  url = {https://arxiv.org/abs/1802.09564},
  
  author = {Zhu, Yuke and Wang, Ziyu and Merel, Josh and Rusu, Andrei and Erez, Tom and Cabi, Serkan and Tunyasuvunakool, Saran and Kramár, János and Hadsell, Raia and de Freitas, Nando and Heess, Nicolas},
  
  title = {Reinforcement and Imitation Learning for Diverse Visuomotor Skills},
  
  publisher = {arXiv},
  
  year = {2018},
  
  copyright = {arXiv.org perpetual, non-exclusive license}
}

@misc{dexterous,
  doi = {10.48550/ARXIV.2203.13251},
  
  url = {https://arxiv.org/abs/2203.13251},
  
  author = {Arunachalam, Sridhar Pandian and Silwal, Sneha and Evans, Ben and Pinto, Lerrel},
  
  title = {Dexterous Imitation Made Easy: A Learning-Based Framework for Efficient Dexterous Manipulation},
  
  publisher = {arXiv},
  
  year = {2022},
  
  copyright = {arXiv.org perpetual, non-exclusive license}
}

@article{2014Reinforcement,
  title={Reinforcement Learning in Robotics: A Survey},
  author={ Bagnell, J. A. },
  year={2014},
}

@misc{survey,
  doi = {10.48550/ARXIV.1803.10862},
  
  url = {https://arxiv.org/abs/1803.10862},
  
  author = {Ruiz-del-Solar, Javier and Loncomilla, Patricio and Soto, Naiomi},
  
  title = {A Survey on Deep Learning Methods for Robot Vision},
  
  publisher = {arXiv},
  
  year = {2018},
  
  copyright = {arXiv.org perpetual, non-exclusive license}
}

@misc{vision2022,
  doi = {10.48550/ARXIV.2203.12677},
  
  url = {https://arxiv.org/abs/2203.12677},
  
  author = {Hsu, Kyle and Kim, Moo Jin and Rafailov, Rafael and Wu, Jiajun and Finn, Chelsea},
  
  title = {Vision-Based Manipulators Need to Also See from Their Hands},
  
  publisher = {arXiv},
  
  year = {2022},
  
  copyright = {Creative Commons Attribution 4.0 International}
}

@article{ROOPAK1993SIGNATURE,
  title={SIGNATURE VERIFICATION USING A "SIAMESE" TIME DELAY NEURAL NETWORK},
  author={ROOPAK and SHAH and EDUARD and SCKINGER and JAMES and W. and BENTZ and ISABELLE and GUYON and CLIFF and },
  journal={International Journal of Pattern Recognition and Artificial Intelligence},
  volume={07},
  number={4},
  pages={669-669},
  year={1993},
}

@misc{unet,
  doi = {10.48550/ARXIV.1505.04597},
  
  url = {https://arxiv.org/abs/1505.04597},
  
  author = {Ronneberger, Olaf and Fischer, Philipp and Brox, Thomas},
  
  title = {U-Net: Convolutional Networks for Biomedical Image Segmentation},
  
  publisher = {arXiv},
  
  year = {2015},
  
  copyright = {arXiv.org perpetual, non-exclusive license}
}

@inproceedings{BMVC2016_119,
        	title={Learning local feature descriptors with triplets and shallow convolutional neural networks},
        	author={Vassileios Balntas, Edgar Riba, Daniel Ponsa and Krystian  Mikolajczyk},
        	year={2016},
        	month={September},
        	pages={119.1-119.11},
        	articleno={119},
        	numpages={11},
        	booktitle={Proceedings of the British Machine Vision Conference (BMVC)},
        	publisher={BMVA Press},
        	editor={Richard C. Wilson, Edwin R. Hancock and William A. P. Smith},
        	doi={10.5244/C.30.119},
        	isbn={1-901725-59-6},
        	url={https://dx.doi.org/10.5244/C.30.119}
}

@misc{pyrender,
author = {Matthew Matl},
title = {Pyrender},
year = {2019},
publisher = {GitHub},
journal = {GitHub repository},
howpublished = {\url{https://github.com/mmatl/pyrender}}
}

@article{erwin2019Python,
  title={PyBullet, a Python module for physics simulation for games},
  author={Erwin Coumans and Yunfei Bai},
  journal={robotics and machine learning},
  year={2016-2019},
  url={http://pybullet.org}
}

@article{Pan2010ASO,
  title={A Survey on Transfer Learning},
  author={Sinno Jialin Pan and Qiang Yang},
  journal={IEEE Transactions on Knowledge and Data Engineering},
  year={2010},
  volume={22},
  pages={1345-1359}
}

@misc{yosinski2014transferable,
      title={How transferable are features in deep neural networks?}, 
      author={Jason Yosinski and Jeff Clune and Yoshua Bengio and Hod Lipson},
      year={2014},
      eprint={1411.1792},
      archivePrefix={arXiv},
      primaryClass={cs.LG}
}

@INPROCEEDINGS{5206848,  author={Deng, Jia and Dong, Wei and Socher, Richard and Li, Li-Jia and Kai Li and Li Fei-Fei},  booktitle={2009 IEEE Conference on Computer Vision and Pattern Recognition},   title={ImageNet: A large-scale hierarchical image database},   year={2009},  volume={},  number={},  pages={248-255},  doi={10.1109/CVPR.2009.5206848}}

@InProceedings{pmlr-v27-bengio12a,
  title = 	 {Deep Learning of Representations for Unsupervised and Transfer Learning},
  author = 	 {Bengio, Yoshua},
  booktitle = 	 {Proceedings of ICML Workshop on Unsupervised and Transfer Learning},
  pages = 	 {17--36},
  year = 	 {2012},
  editor = 	 {Guyon, Isabelle and Dror, Gideon and Lemaire, Vincent and Taylor, Graham and Silver, Daniel},
  volume = 	 {27},
  series = 	 {Proceedings of Machine Learning Research},
  address = 	 {Bellevue, Washington, USA},
  month = 	 {02 Jul},
  publisher =    {PMLR},
  url = 	 {https://proceedings.mlr.press/v27/bengio12a.html},
}

@misc{mikolov2013efficient,
      title={Efficient Estimation of Word Representations in Vector Space}, 
      author={Tomas Mikolov and Kai Chen and Greg Corrado and Jeffrey Dean},
      year={2013},
      eprint={1301.3781},
      archivePrefix={arXiv},
      primaryClass={cs.CL}
}

@inproceedings{pennington2014glove,
  author = {Jeffrey Pennington and Richard Socher and Christopher D. Manning},
  booktitle = {Empirical Methods in Natural Language Processing (EMNLP)},
  title = {GloVe: Global Vectors for Word Representation},
  year = {2014},
  pages = {1532--1543},
  url = {http://www.aclweb.org/anthology/D14-1162},
}

@misc{du2021fewshot,
      title={Few-Shot Learning via Learning the Representation, Provably},
      author={Simon S. Du and Wei Hu and Sham M. Kakade and Jason D. Lee and Qi Lei},
      year={2021},
      eprint={2002.09434},
      archivePrefix={arXiv},
      primaryClass={cs.LG}
}

@misc{danielczuk2019segmenting,
      title={Segmenting Unknown 3D Objects from Real Depth Images using Mask R-CNN Trained on Synthetic Data}, 
      author={Michael Danielczuk and Matthew Matl and Saurabh Gupta and Andrew Li and Andrew Lee and Jeffrey Mahler and Ken Goldberg},
      year={2019},
      eprint={1809.05825},
      archivePrefix={arXiv},
      primaryClass={cs.CV}
}

@misc{mahler2018dexnet,
      title={Dex-Net 3.0: Computing Robust Robot Vacuum Suction Grasp Targets in Point Clouds using a New Analytic Model and Deep Learning}, 
      author={Jeffrey Mahler and Matthew Matl and Xinyu Liu and Albert Li and David Gealy and Ken Goldberg},
      year={2018},
      eprint={1709.06670},
      archivePrefix={arXiv},
      primaryClass={cs.RO}
}

@misc{openai2019solving,
      title={Solving Rubik's Cube with a Robot Hand}, 
      author={OpenAI and Ilge Akkaya and Marcin Andrychowicz and Maciek Chociej and Mateusz Litwin and Bob McGrew and Arthur Petron and Alex Paino and Matthias Plappert and Glenn Powell and Raphael Ribas and Jonas Schneider and Nikolas Tezak and Jerry Tworek and Peter Welinder and Lilian Weng and Qiming Yuan and Wojciech Zaremba and Lei Zhang},
      year={2019},
      eprint={1910.07113},
      archivePrefix={arXiv},
      primaryClass={cs.LG}
}

@article{Tzeng2015TowardsAD,
  title={Towards Adapting Deep Visuomotor Representations from Simulated to Real Environments},
  author={Eric Tzeng and Coline Devin and Judy Hoffman and Chelsea Finn and Xingchao Peng and Sergey Levine and Kate Saenko and Trevor Darrell},
  journal={ArXiv},
  year={2015},
  volume={abs/1511.07111}
}

@misc{christiano2016transfer,
      title={Transfer from Simulation to Real World through Learning Deep Inverse Dynamics Model}, 
      author={Paul Christiano and Zain Shah and Igor Mordatch and Jonas Schneider and Trevor Blackwell and Joshua Tobin and Pieter Abbeel and Wojciech Zaremba},
      year={2016},
      eprint={1610.03518},
      archivePrefix={arXiv},
      primaryClass={cs.RO}
}

@misc{rusu2018simtoreal,
      title={Sim-to-Real Robot Learning from Pixels with Progressive Nets}, 
      author={Andrei A. Rusu and Mel Vecerik and Thomas Rothörl and Nicolas Heess and Razvan Pascanu and Raia Hadsell},
      year={2018},
      eprint={1610.04286},
      archivePrefix={arXiv},
      primaryClass={cs.RO}
}

\end{document}